\documentclass[sigconf,nonacm]{aamas}

\usepackage{balance}
\usepackage{booktabs}
\usepackage{amsmath}
\usepackage{siunitx}
\newcommand{\rd}[1]{\num[round-mode=places,round-precision=2,retain-explicit-plus=true]{#1}}
\usepackage{graphicx}
\graphicspath{{figures/}}
\usepackage{tikz}
\usetikzlibrary{positioning,calc,fit,arrows.meta,backgrounds}

\def\spPos{0.00/0.00,4.20/0.00,8.40/0.00,46.80/0.00,93.60/0.00,0.00/-2.55,4.20/-2.55,8.40/-2.55,46.80/-2.55,93.60/-2.55,0.00/-5.10,4.20/-5.10,8.40/-5.10,46.80/-5.10,93.60/-5.10,97.80/-5.10,0.00/-7.65,4.20/-7.65,8.40/-7.65,93.60/-7.65,97.80/-7.65,0.00/-10.20,4.20/-10.20,46.80/-10.20,51.00/-10.20,0.00/-12.75,4.20/-12.75,93.60/-12.75,0.00/-15.30,4.20/-15.30,93.60/-15.30,97.80/-15.30,0.00/-17.85,4.20/-17.85,0.00/-20.40,4.20/-20.40,0.00/-22.95,4.20/-22.95,93.60/-22.95,0.00/-25.50,4.20/-25.50,93.60/-25.50,97.80/-25.50,102.00/-25.50,0.00/-28.05,4.20/-28.05,93.60/-28.05,97.80/-28.05,102.00/-28.05,0.00/-30.60,4.20/-30.60,93.60/-30.60,97.80/-30.60,0.00/-33.15,4.20/-33.15,93.60/-33.15,0.00/-35.70,46.80/-35.70,93.60/-35.70,0.00/-38.25,0.00/-40.80,93.60/-40.80,0.00/-43.35,0.00/-45.90,93.60/-45.90,97.80/-45.90,0.00/-48.45,93.60/-48.45,97.80/-48.45,0.00/-51.00,93.60/-51.00,0.00/-53.55,93.60/-53.55,0.00/-56.10,93.60/-56.10,97.80/-56.10,102.00/-56.10,0.00/-58.65,93.60/-58.65,0.00/-61.20,93.60/-61.20,97.80/-61.20,0.00/-63.75,46.80/-63.75,93.60/-63.75,0.00/-66.30,0.00/-68.85,93.60/-68.85,0.00/-71.40,93.60/-71.40,97.80/-71.40,0.00/-73.95,0.00/-76.50,46.80/-76.50,93.60/-76.50,97.80/-76.50,102.00/-76.50,106.20/-76.50,0.00/-79.05,93.60/-79.05,97.80/-79.05,93.60/-81.60,93.60/-84.15,97.80/-84.15,93.60/-86.70,93.60/-89.25,97.80/-89.25,93.60/-94.35,97.80/-94.35,46.80/-96.90,93.60/-96.90,97.80/-96.90,102.00/-96.90,93.60/-99.45,93.60/-102.00,93.60/-104.55,97.80/-104.55,93.60/-107.10,93.60/-109.65,97.80/-109.65,93.60/-112.20,93.60/-114.75,93.60/-117.30,97.80/-117.30,93.60/-119.85,93.60/-122.40,93.60/-124.95,93.60/-127.50,93.60/-130.05,97.80/-130.05,102.00/-130.05,46.80/-132.60,93.60/-135.15,93.60/-137.70,93.60/-140.25,97.80/-140.25,93.60/-142.80,93.60/-145.35,93.60/-147.90,93.60/-150.45,97.80/-150.45,102.00/-150.45}
\def\spNeg{59.40/-7.65,152.40/-7.65,80.40/-10.20,139.80/-10.20,67.80/-15.30,148.20/-15.30,152.40/-17.85,148.20/-28.05,148.20/-38.25,152.40/-38.25,144.00/-45.90,148.20/-45.90,152.40/-45.90,152.40/-48.45,148.20/-51.00,46.80/-53.55,144.00/-61.20,148.20/-61.20,148.20/-66.30,152.40/-66.30,152.40/-71.40,152.40/-79.05,12.60/-81.60,55.20/-84.15,55.20/-86.70,4.20/-99.45,152.40/-114.75,148.20/-117.30,152.40/-122.40,152.40/-124.95,152.40/-130.05,152.40/-137.70,152.40/-142.80,152.40/-150.45}
\def\spZero{12.60/0.00,16.80/0.00,21.00/0.00,25.20/0.00,29.40/0.00,33.60/0.00,51.00/0.00,55.20/0.00,59.40/0.00,63.60/0.00,67.80/0.00,72.00/0.00,76.20/0.00,80.40/0.00,97.80/0.00,102.00/0.00,106.20/0.00,110.40/0.00,114.60/0.00,118.80/0.00,123.00/0.00,127.20/0.00,131.40/0.00,135.60/0.00,139.80/0.00,144.00/0.00,148.20/0.00,12.60/-2.55,16.80/-2.55,21.00/-2.55,25.20/-2.55,29.40/-2.55,51.00/-2.55,55.20/-2.55,59.40/-2.55,63.60/-2.55,67.80/-2.55,72.00/-2.55,76.20/-2.55,80.40/-2.55,97.80/-2.55,102.00/-2.55,106.20/-2.55,110.40/-2.55,114.60/-2.55,118.80/-2.55,123.00/-2.55,127.20/-2.55,131.40/-2.55,135.60/-2.55,139.80/-2.55,12.60/-5.10,51.00/-5.10,55.20/-5.10,59.40/-5.10,102.00/-5.10,106.20/-5.10,110.40/-5.10,114.60/-5.10,118.80/-5.10,123.00/-5.10,127.20/-5.10,131.40/-5.10,135.60/-5.10,139.80/-5.10,144.00/-5.10,148.20/-5.10,152.40/-5.10,46.80/-7.65,51.00/-7.65,55.20/-7.65,102.00/-7.65,106.20/-7.65,110.40/-7.65,114.60/-7.65,118.80/-7.65,123.00/-7.65,127.20/-7.65,131.40/-7.65,135.60/-7.65,139.80/-7.65,144.00/-7.65,148.20/-7.65,8.40/-10.20,12.60/-10.20,16.80/-10.20,21.00/-10.20,25.20/-10.20,29.40/-10.20,55.20/-10.20,59.40/-10.20,63.60/-10.20,67.80/-10.20,72.00/-10.20,76.20/-10.20,93.60/-10.20,97.80/-10.20,102.00/-10.20,106.20/-10.20,110.40/-10.20,114.60/-10.20,118.80/-10.20,123.00/-10.20,127.20/-10.20,131.40/-10.20,135.60/-10.20,8.40/-12.75,12.60/-12.75,16.80/-12.75,46.80/-12.75,51.00/-12.75,55.20/-12.75,59.40/-12.75,63.60/-12.75,67.80/-12.75,97.80/-12.75,102.00/-12.75,106.20/-12.75,110.40/-12.75,114.60/-12.75,118.80/-12.75,123.00/-12.75,127.20/-12.75,131.40/-12.75,135.60/-12.75,139.80/-12.75,144.00/-12.75,148.20/-12.75,8.40/-15.30,12.60/-15.30,16.80/-15.30,46.80/-15.30,51.00/-15.30,55.20/-15.30,59.40/-15.30,63.60/-15.30,102.00/-15.30,106.20/-15.30,110.40/-15.30,114.60/-15.30,118.80/-15.30,123.00/-15.30,127.20/-15.30,131.40/-15.30,135.60/-15.30,139.80/-15.30,144.00/-15.30,8.40/-17.85,46.80/-17.85,51.00/-17.85,55.20/-17.85,59.40/-17.85,93.60/-17.85,97.80/-17.85,102.00/-17.85,106.20/-17.85,110.40/-17.85,114.60/-17.85,118.80/-17.85,123.00/-17.85,127.20/-17.85,131.40/-17.85,135.60/-17.85,139.80/-17.85,144.00/-17.85,148.20/-17.85,8.40/-20.40,46.80/-20.40,51.00/-20.40,55.20/-20.40,93.60/-20.40,97.80/-20.40,102.00/-20.40,106.20/-20.40,110.40/-20.40,114.60/-20.40,118.80/-20.40,123.00/-20.40,127.20/-20.40,131.40/-20.40,135.60/-20.40,139.80/-20.40,144.00/-20.40,148.20/-20.40,152.40/-20.40,8.40/-22.95,46.80/-22.95,51.00/-22.95,55.20/-22.95,97.80/-22.95,102.00/-22.95,106.20/-22.95,110.40/-22.95,114.60/-22.95,118.80/-22.95,123.00/-22.95,127.20/-22.95,131.40/-22.95,135.60/-22.95,139.80/-22.95,144.00/-22.95,148.20/-22.95,152.40/-22.95,8.40/-25.50,46.80/-25.50,51.00/-25.50,55.20/-25.50,106.20/-25.50,110.40/-25.50,114.60/-25.50,118.80/-25.50,123.00/-25.50,127.20/-25.50,131.40/-25.50,135.60/-25.50,139.80/-25.50,144.00/-25.50,148.20/-25.50,8.40/-28.05,46.80/-28.05,51.00/-28.05,55.20/-28.05,106.20/-28.05,110.40/-28.05,114.60/-28.05,118.80/-28.05,123.00/-28.05,127.20/-28.05,131.40/-28.05,135.60/-28.05,139.80/-28.05,144.00/-28.05,46.80/-30.60,51.00/-30.60,102.00/-30.60,106.20/-30.60,110.40/-30.60,114.60/-30.60,118.80/-30.60,123.00/-30.60,127.20/-30.60,131.40/-30.60,135.60/-30.60,139.80/-30.60,144.00/-30.60,148.20/-30.60,152.40/-30.60,46.80/-33.15,51.00/-33.15,97.80/-33.15,102.00/-33.15,106.20/-33.15,110.40/-33.15,114.60/-33.15,118.80/-33.15,123.00/-33.15,127.20/-33.15,131.40/-33.15,135.60/-33.15,139.80/-33.15,144.00/-33.15,148.20/-33.15,4.20/-35.70,8.40/-35.70,12.60/-35.70,51.00/-35.70,55.20/-35.70,59.40/-35.70,97.80/-35.70,102.00/-35.70,106.20/-35.70,110.40/-35.70,114.60/-35.70,118.80/-35.70,123.00/-35.70,127.20/-35.70,131.40/-35.70,135.60/-35.70,139.80/-35.70,144.00/-35.70,4.20/-38.25,8.40/-38.25,46.80/-38.25,51.00/-38.25,55.20/-38.25,59.40/-38.25,93.60/-38.25,97.80/-38.25,102.00/-38.25,106.20/-38.25,110.40/-38.25,114.60/-38.25,118.80/-38.25,123.00/-38.25,127.20/-38.25,131.40/-38.25,135.60/-38.25,139.80/-38.25,144.00/-38.25,4.20/-40.80,46.80/-40.80,51.00/-40.80,97.80/-40.80,102.00/-40.80,106.20/-40.80,110.40/-40.80,114.60/-40.80,118.80/-40.80,123.00/-40.80,127.20/-40.80,131.40/-40.80,135.60/-40.80,139.80/-40.80,144.00/-40.80,148.20/-40.80,4.20/-43.35,46.80/-43.35,51.00/-43.35,93.60/-43.35,97.80/-43.35,102.00/-43.35,106.20/-43.35,110.40/-43.35,114.60/-43.35,118.80/-43.35,123.00/-43.35,127.20/-43.35,131.40/-43.35,135.60/-43.35,139.80/-43.35,144.00/-43.35,148.20/-43.35,4.20/-45.90,46.80/-45.90,51.00/-45.90,102.00/-45.90,106.20/-45.90,110.40/-45.90,114.60/-45.90,118.80/-45.90,123.00/-45.90,127.20/-45.90,131.40/-45.90,135.60/-45.90,139.80/-45.90,4.20/-48.45,46.80/-48.45,51.00/-48.45,102.00/-48.45,106.20/-48.45,110.40/-48.45,114.60/-48.45,118.80/-48.45,123.00/-48.45,127.20/-48.45,131.40/-48.45,135.60/-48.45,139.80/-48.45,144.00/-48.45,148.20/-48.45,4.20/-51.00,46.80/-51.00,51.00/-51.00,97.80/-51.00,102.00/-51.00,106.20/-51.00,110.40/-51.00,114.60/-51.00,118.80/-51.00,123.00/-51.00,127.20/-51.00,131.40/-51.00,135.60/-51.00,139.80/-51.00,144.00/-51.00,97.80/-53.55,102.00/-53.55,106.20/-53.55,110.40/-53.55,114.60/-53.55,118.80/-53.55,123.00/-53.55,127.20/-53.55,131.40/-53.55,135.60/-53.55,139.80/-53.55,144.00/-53.55,148.20/-53.55,152.40/-53.55,46.80/-56.10,106.20/-56.10,110.40/-56.10,114.60/-56.10,118.80/-56.10,123.00/-56.10,127.20/-56.10,131.40/-56.10,135.60/-56.10,139.80/-56.10,144.00/-56.10,148.20/-56.10,152.40/-56.10,46.80/-58.65,97.80/-58.65,102.00/-58.65,106.20/-58.65,110.40/-58.65,114.60/-58.65,118.80/-58.65,123.00/-58.65,127.20/-58.65,131.40/-58.65,135.60/-58.65,139.80/-58.65,144.00/-58.65,148.20/-58.65,152.40/-58.65,46.80/-61.20,102.00/-61.20,106.20/-61.20,110.40/-61.20,114.60/-61.20,118.80/-61.20,123.00/-61.20,127.20/-61.20,131.40/-61.20,135.60/-61.20,139.80/-61.20,97.80/-63.75,102.00/-63.75,106.20/-63.75,110.40/-63.75,114.60/-63.75,118.80/-63.75,123.00/-63.75,127.20/-63.75,131.40/-63.75,135.60/-63.75,139.80/-63.75,144.00/-63.75,148.20/-63.75,46.80/-66.30,93.60/-66.30,97.80/-66.30,102.00/-66.30,106.20/-66.30,110.40/-66.30,114.60/-66.30,118.80/-66.30,123.00/-66.30,127.20/-66.30,131.40/-66.30,135.60/-66.30,139.80/-66.30,144.00/-66.30,46.80/-68.85,97.80/-68.85,102.00/-68.85,106.20/-68.85,110.40/-68.85,114.60/-68.85,118.80/-68.85,123.00/-68.85,127.20/-68.85,131.40/-68.85,135.60/-68.85,139.80/-68.85,144.00/-68.85,148.20/-68.85,152.40/-68.85,46.80/-71.40,102.00/-71.40,106.20/-71.40,110.40/-71.40,114.60/-71.40,118.80/-71.40,123.00/-71.40,127.20/-71.40,131.40/-71.40,135.60/-71.40,139.80/-71.40,144.00/-71.40,148.20/-71.40,46.80/-73.95,93.60/-73.95,97.80/-73.95,102.00/-73.95,106.20/-73.95,110.40/-73.95,114.60/-73.95,118.80/-73.95,123.00/-73.95,127.20/-73.95,131.40/-73.95,135.60/-73.95,139.80/-73.95,144.00/-73.95,148.20/-73.95,110.40/-76.50,114.60/-76.50,118.80/-76.50,123.00/-76.50,127.20/-76.50,131.40/-76.50,135.60/-76.50,139.80/-76.50,144.00/-76.50,148.20/-76.50,152.40/-76.50,46.80/-79.05,102.00/-79.05,106.20/-79.05,110.40/-79.05,114.60/-79.05,118.80/-79.05,123.00/-79.05,127.20/-79.05,131.40/-79.05,135.60/-79.05,139.80/-79.05,144.00/-79.05,148.20/-79.05,0.00/-81.60,4.20/-81.60,8.40/-81.60,46.80/-81.60,51.00/-81.60,55.20/-81.60,59.40/-81.60,97.80/-81.60,102.00/-81.60,106.20/-81.60,110.40/-81.60,114.60/-81.60,118.80/-81.60,123.00/-81.60,127.20/-81.60,131.40/-81.60,135.60/-81.60,139.80/-81.60,144.00/-81.60,148.20/-81.60,152.40/-81.60,0.00/-84.15,4.20/-84.15,8.40/-84.15,46.80/-84.15,51.00/-84.15,102.00/-84.15,106.20/-84.15,110.40/-84.15,114.60/-84.15,118.80/-84.15,123.00/-84.15,127.20/-84.15,131.40/-84.15,135.60/-84.15,139.80/-84.15,144.00/-84.15,148.20/-84.15,0.00/-86.70,4.20/-86.70,8.40/-86.70,46.80/-86.70,51.00/-86.70,97.80/-86.70,102.00/-86.70,106.20/-86.70,110.40/-86.70,114.60/-86.70,118.80/-86.70,123.00/-86.70,127.20/-86.70,131.40/-86.70,135.60/-86.70,139.80/-86.70,144.00/-86.70,148.20/-86.70,152.40/-86.70,0.00/-89.25,4.20/-89.25,8.40/-89.25,46.80/-89.25,51.00/-89.25,55.20/-89.25,102.00/-89.25,106.20/-89.25,110.40/-89.25,114.60/-89.25,118.80/-89.25,123.00/-89.25,127.20/-89.25,131.40/-89.25,135.60/-89.25,139.80/-89.25,144.00/-89.25,148.20/-89.25,0.00/-91.80,4.20/-91.80,8.40/-91.80,46.80/-91.80,51.00/-91.80,55.20/-91.80,93.60/-91.80,97.80/-91.80,102.00/-91.80,106.20/-91.80,110.40/-91.80,114.60/-91.80,118.80/-91.80,123.00/-91.80,127.20/-91.80,131.40/-91.80,135.60/-91.80,139.80/-91.80,144.00/-91.80,148.20/-91.80,152.40/-91.80,0.00/-94.35,4.20/-94.35,8.40/-94.35,46.80/-94.35,51.00/-94.35,55.20/-94.35,102.00/-94.35,106.20/-94.35,110.40/-94.35,114.60/-94.35,118.80/-94.35,123.00/-94.35,127.20/-94.35,131.40/-94.35,135.60/-94.35,139.80/-94.35,144.00/-94.35,148.20/-94.35,152.40/-94.35,0.00/-96.90,4.20/-96.90,8.40/-96.90,51.00/-96.90,55.20/-96.90,106.20/-96.90,110.40/-96.90,114.60/-96.90,118.80/-96.90,123.00/-96.90,127.20/-96.90,131.40/-96.90,135.60/-96.90,139.80/-96.90,144.00/-96.90,148.20/-96.90,0.00/-99.45,46.80/-99.45,51.00/-99.45,97.80/-99.45,102.00/-99.45,106.20/-99.45,110.40/-99.45,114.60/-99.45,118.80/-99.45,123.00/-99.45,127.20/-99.45,131.40/-99.45,135.60/-99.45,139.80/-99.45,144.00/-99.45,0.00/-102.00,4.20/-102.00,46.80/-102.00,51.00/-102.00,97.80/-102.00,102.00/-102.00,106.20/-102.00,110.40/-102.00,114.60/-102.00,118.80/-102.00,123.00/-102.00,127.20/-102.00,131.40/-102.00,135.60/-102.00,139.80/-102.00,144.00/-102.00,148.20/-102.00,152.40/-102.00,0.00/-104.55,4.20/-104.55,46.80/-104.55,51.00/-104.55,102.00/-104.55,106.20/-104.55,110.40/-104.55,114.60/-104.55,118.80/-104.55,123.00/-104.55,127.20/-104.55,131.40/-104.55,135.60/-104.55,139.80/-104.55,144.00/-104.55,148.20/-104.55,0.00/-107.10,4.20/-107.10,46.80/-107.10,51.00/-107.10,97.80/-107.10,102.00/-107.10,106.20/-107.10,110.40/-107.10,114.60/-107.10,118.80/-107.10,123.00/-107.10,127.20/-107.10,131.40/-107.10,135.60/-107.10,139.80/-107.10,144.00/-107.10,148.20/-107.10,0.00/-109.65,4.20/-109.65,46.80/-109.65,51.00/-109.65,102.00/-109.65,106.20/-109.65,110.40/-109.65,114.60/-109.65,118.80/-109.65,123.00/-109.65,127.20/-109.65,131.40/-109.65,135.60/-109.65,139.80/-109.65,144.00/-109.65,148.20/-109.65,152.40/-109.65,0.00/-112.20,4.20/-112.20,46.80/-112.20,51.00/-112.20,97.80/-112.20,102.00/-112.20,106.20/-112.20,110.40/-112.20,114.60/-112.20,118.80/-112.20,123.00/-112.20,127.20/-112.20,131.40/-112.20,135.60/-112.20,139.80/-112.20,144.00/-112.20,148.20/-112.20,152.40/-112.20,0.00/-114.75,4.20/-114.75,46.80/-114.75,51.00/-114.75,97.80/-114.75,102.00/-114.75,106.20/-114.75,110.40/-114.75,114.60/-114.75,118.80/-114.75,123.00/-114.75,127.20/-114.75,131.40/-114.75,135.60/-114.75,139.80/-114.75,144.00/-114.75,148.20/-114.75,0.00/-117.30,46.80/-117.30,102.00/-117.30,106.20/-117.30,110.40/-117.30,114.60/-117.30,118.80/-117.30,123.00/-117.30,127.20/-117.30,131.40/-117.30,135.60/-117.30,139.80/-117.30,144.00/-117.30,0.00/-119.85,46.80/-119.85,97.80/-119.85,102.00/-119.85,106.20/-119.85,110.40/-119.85,114.60/-119.85,118.80/-119.85,123.00/-119.85,127.20/-119.85,131.40/-119.85,135.60/-119.85,139.80/-119.85,144.00/-119.85,148.20/-119.85,152.40/-119.85,0.00/-122.40,46.80/-122.40,97.80/-122.40,102.00/-122.40,106.20/-122.40,110.40/-122.40,114.60/-122.40,118.80/-122.40,123.00/-122.40,127.20/-122.40,131.40/-122.40,135.60/-122.40,139.80/-122.40,144.00/-122.40,148.20/-122.40,0.00/-124.95,46.80/-124.95,97.80/-124.95,102.00/-124.95,106.20/-124.95,110.40/-124.95,114.60/-124.95,118.80/-124.95,123.00/-124.95,127.20/-124.95,131.40/-124.95,135.60/-124.95,139.80/-124.95,144.00/-124.95,148.20/-124.95,0.00/-127.50,46.80/-127.50,97.80/-127.50,102.00/-127.50,106.20/-127.50,110.40/-127.50,114.60/-127.50,118.80/-127.50,123.00/-127.50,127.20/-127.50,131.40/-127.50,135.60/-127.50,139.80/-127.50,144.00/-127.50,148.20/-127.50,152.40/-127.50,0.00/-130.05,46.80/-130.05,106.20/-130.05,110.40/-130.05,114.60/-130.05,118.80/-130.05,123.00/-130.05,127.20/-130.05,131.40/-130.05,135.60/-130.05,139.80/-130.05,144.00/-130.05,148.20/-130.05,0.00/-132.60,93.60/-132.60,97.80/-132.60,102.00/-132.60,106.20/-132.60,110.40/-132.60,114.60/-132.60,118.80/-132.60,123.00/-132.60,127.20/-132.60,131.40/-132.60,135.60/-132.60,139.80/-132.60,144.00/-132.60,148.20/-132.60,152.40/-132.60,0.00/-135.15,46.80/-135.15,97.80/-135.15,102.00/-135.15,106.20/-135.15,110.40/-135.15,114.60/-135.15,118.80/-135.15,123.00/-135.15,127.20/-135.15,131.40/-135.15,135.60/-135.15,139.80/-135.15,144.00/-135.15,148.20/-135.15,152.40/-135.15,0.00/-137.70,46.80/-137.70,97.80/-137.70,102.00/-137.70,106.20/-137.70,110.40/-137.70,114.60/-137.70,118.80/-137.70,123.00/-137.70,127.20/-137.70,131.40/-137.70,135.60/-137.70,139.80/-137.70,144.00/-137.70,148.20/-137.70,0.00/-140.25,46.80/-140.25,102.00/-140.25,106.20/-140.25,110.40/-140.25,114.60/-140.25,118.80/-140.25,123.00/-140.25,127.20/-140.25,131.40/-140.25,135.60/-140.25,139.80/-140.25,144.00/-140.25,148.20/-140.25,152.40/-140.25,0.00/-142.80,46.80/-142.80,97.80/-142.80,102.00/-142.80,106.20/-142.80,110.40/-142.80,114.60/-142.80,118.80/-142.80,123.00/-142.80,127.20/-142.80,131.40/-142.80,135.60/-142.80,139.80/-142.80,144.00/-142.80,148.20/-142.80,0.00/-145.35,46.80/-145.35,97.80/-145.35,102.00/-145.35,106.20/-145.35,110.40/-145.35,114.60/-145.35,118.80/-145.35,123.00/-145.35,127.20/-145.35,131.40/-145.35,135.60/-145.35,139.80/-145.35,144.00/-145.35,148.20/-145.35,152.40/-145.35,0.00/-147.90,46.80/-147.90,97.80/-147.90,102.00/-147.90,106.20/-147.90,110.40/-147.90,114.60/-147.90,118.80/-147.90,123.00/-147.90,127.20/-147.90,131.40/-147.90,135.60/-147.90,139.80/-147.90,144.00/-147.90,148.20/-147.90,152.40/-147.90,0.00/-150.45,46.80/-150.45,106.20/-150.45,110.40/-150.45,114.60/-150.45,118.80/-150.45,123.00/-150.45,127.20/-150.45,131.40/-150.45,135.60/-150.45,139.80/-150.45,144.00/-150.45,148.20/-150.45}

\def\spTotH{153.00pt}

\def\spBearMid{18.90pt}
\def\spBearN{139}
\def\spBearPos{50}

\def\spBearRate{36.0}
\def\spBearLo{28.5}
\def\spBearHi{44.2}

\def\spCtrlMid{65.70pt}
\def\spCtrlN{146}
\def\spCtrlPos{10}

\def\spCtrlRate{6.8}
\def\spCtrlLo{3.8}
\def\spCtrlHi{12.1}

\def\spRandMid{125.10pt}
\def\spRandN{872}
\def\spRandPos{82}

\def\spRandRate{9.4}
\def\spRandLo{7.6}
\def\spRandHi{11.5}
\def\spCells{1{,}157}
\def\spMems{60}
\def\spNonzero{176}
\def\spZeroPct{84.8}

\def\spBearTasks{56}

\def\spCtrlTasks{100}

\def\spRandTasks{147}

\def\bearUnits{1{,}889}

\def\bearDetTasks{75}

\def\stratBearV{+0.3453}

\def\stratCtrlV{+0.0274}

\def\stratRandV{+0.0642}

\def\simRhoAll{+0.087}
\def\simRhoRand{+0.006}
\def\simCells{1{,}157}

\def\frRawDU{+0.3174}
\def\frRawLo{+0.192}
\def\frRawHi{+0.448}

\def\frRawTokExp{9{,}228}

\def\frSumDU{+0.0613}
\def\frSumLo{-0.042}
\def\frSumHi{+0.164}

\def\frSumTokExp{323}

\def\frStrDU{-0.0498}
\def\frStrLo{-0.143}
\def\frStrHi{+0.038}

\def\frStrTokExp{238}

\def\frCells{277}
\def\frMems{57}
\def\frReuse{4.86}

\def\xbSRLmM{-0.1408}
\def\xbSRLmLo{-0.2211}
\def\xbSRLmHi{-0.0601}
\def\xbSRLmN{277}
\def\xbSRScM{+0.0246}
\def\xbSRScLo{-0.0342}
\def\xbSRScHi{+0.0991}
\def\xbSRScN{122}
\def\xbSRSwM{-0.1000}
\def\xbSRSwLo{-0.2250}
\def\xbSRSwHi{+0.0000}
\def\xbSRSwN{40}
\def\xbTRLmM{-0.1625}
\def\xbTRLmLo{-0.2454}
\def\xbTRLmHi{-0.0798}
\def\xbTRLmN{277}
\def\xbTRScM{+0.0328}
\def\xbTRScLo{-0.0342}
\def\xbTRScHi{+0.1200}
\def\xbTRScN{122}
\def\xbTRSwM{-0.0500}
\def\xbTRSwLo{-0.1750}
\def\xbTRSwHi{+0.0750}
\def\xbTRSwN{40}

\def\sciPairs{154}
\def\sciMems{34}
\def\sciTasks{40}

\def\sweMatched{40}
\def\sweLocMiss{38}

\def\sweDomRepoPct{92}
\def\rgSmDiff{+0.0982}
\def\rgSmLo{+0.0247}
\def\rgSmHi{+0.1730}
\def\rgSmN{550}
\def\rgSmCells{275}
\def\rgSmTasks{123}

\def\lcPrimary{+0.0519}
\def\lcPrimaryLo{-0.0678}
\def\lcPrimaryHi{+0.1727}
\def\lcEnds{+0.1037}
\def\lcEndsLo{+0.0091}
\def\lcEndsHi{+0.2098}
\def\lcBudget{+0.2612}
\def\lcBudgetLo{+0.1533}
\def\lcBudgetHi{+0.3864}
\def\lcPrimaryN{270}
\def\gfWithoutPct{15.6}
\def\gfWithPct{50.0}
\def\gfN{64}
\def\gfCtlWithoutPct{14.3}
\def\gfCtlWithPct{11.9}
\def\gfCtlN{84}
\def\gfDiff{+0.3676}
\def\gfCells{32}
\def\gfLexBaf{0.744}
\def\gfLlmBaf{0.310}

\def\rgSmOss{+0.2182}
\def\rgSmGem{+0.3164}
\def\xbFloorLm{0.025}
\def\xbFloorLmK{7}
\def\xbFloorLmN{285}
\def\xbFloorSc{0.037}
\def\xbFloorScK{5}
\def\xbFloorScN{135}

\def\rgRawN{276}

\def\rgRawDiff{+0.1268}
\def\rgRawLo{+0.0496}
\def\rgRawHi{+0.2050}

\def\rgRawRAll{+0.520}
\def\rgRawRNz{+0.213}
\def\rgRawNNz{109}
\def\rgRawZeroBoth{167}
\def\rgRawSignK{49}
\def\rgRawSignN{50}
\def\rgRawLabK{216}
\def\rgRawLabN{276}

\def\rgSumDiff{+0.0036}
\def\rgSumLo{-0.0484}
\def\rgSumHi{+0.0612}

\def\rgSumRNz{+0.187}
\def\rgSumNNz{36}

\def\rgStrDiff{+0.0112}
\def\rgStrLo{-0.0536}
\def\rgStrHi{+0.0833}

\def\rgStrRNz{-0.070}
\def\rgStrNNz{35}

\def\rgRows{1{,}694}
\def\rgBmcRawOss{+0.3462}
\def\rgBmcRawOssLo{+0.2210}
\def\rgBmcRawOssHi{+0.4761}
\def\rgBmcRawOssMem{59}
\def\rgBmcRawGem{+0.6002}
\def\rgBmcRawGemLo{+0.4787}
\def\rgBmcRawGemHi{+0.7144}

\def\rgWithoutDelta{-0.0688}

\def\smCZeroZero{170}

\def\smAgree{55}
\def\smBoth{58}

\def\smIdentN{570}
\def\smIdentK{570}

\def\covScElig{462}

\def\covScDone{391}
\def\covScExcl{71}

\def\covSwElig{321}

\def\covSwDone{141}
\def\covSwExcl{180}

\def\covSwRepos{2}

\def\raMatchedMean{0.43}
\def\raMatchedN{35}
\def\raMatchedPZero{63}

\def\raCellsPerMemBear{2.3}
\def\raCellsPerMemRand{14.5}
\def\raNegBearK{2}
\def\raNegBearN{52}

\def\raNegRandK{26}
\def\raNegRandN{108}

\def\raAccZeroBear{11.5}
\def\raAccOneBear{46.0}
\def\raHurtOppBear{16}
\def\raHurtRateBear{12.5}
\def\raHurtLoBear{3.5}
\def\raHurtHiBear{36.0}
\def\raHelpOppBear{123}
\def\raHelpRateBear{40.7}
\def\raHelpLoBear{32.4}
\def\raHelpHiBear{49.5}
\def\raBothCorrectBear{14}
\def\raFailWithBearPct{54}
\def\raAccZeroCtrl{14.4}
\def\raAccOneCtrl{17.1}
\def\raHurtOppCtrl{21}
\def\raHurtRateCtrl{28.6}
\def\raHurtLoCtrl{13.8}
\def\raHurtHiCtrl{50.0}
\def\raHelpOppCtrl{125}
\def\raHelpRateCtrl{8.0}
\def\raHelpLoCtrl{4.4}
\def\raHelpHiCtrl{14.1}

\def\raAccZeroRand{20.5}
\def\raAccOneRand{26.9}
\def\raHurtOppRand{179}
\def\raHurtRateRand{14.5}
\def\raHurtLoRand{10.1}
\def\raHurtHiRand{20.4}
\def\raHelpOppRand{693}
\def\raHelpRateRand{11.8}
\def\raHelpLoRand{9.6}
\def\raHelpHiRand{14.5}

\def\raReaderKappa{0.58}
\def\raReaderLabAgree{225}
\def\raReaderMoving{105}
\def\raHurtOss{6}
\def\raHurtGem{2}

\def\raNonBearHelped{92}

\def\raMdeSRLm{0.12}
\def\raMdeSRSc{0.10}
\def\raMdeSRSw{0.16}
\def\raMdeTRLm{0.12}
\def\raMdeTRSc{0.11}
\def\raMdeTRSw{0.18}
\def\raReuseStar{5.5}
\def\raFormSum{7{,}956}
\def\raFormStr{8{,}141}

\def\lcRawM{+0.34}
\def\lcRawLo{+0.21}
\def\lcRawHi{+0.49}
\def\lcRawKept{100}
\def\lcRawTok{8{,}344}
\def\lcRawKM{+0.34}

\def\lcFourM{+0.31}
\def\lcFourLo{+0.19}
\def\lcFourHi{+0.45}
\def\lcFourKept{84}
\def\lcFourTok{4{,}096}
\def\lcFourKM{+0.36}

\def\lcFourLM{+0.05}

\def\lcOneM{+0.19}
\def\lcOneLo{+0.09}
\def\lcOneHi{+0.31}
\def\lcOneKept{48}
\def\lcOneTok{1{,}024}
\def\lcOneKM{+0.37}

\def\lcOneLM{+0.03}

\def\lcMatchM{+0.08}
\def\lcMatchLo{-0.01}
\def\lcMatchHi{+0.18}
\def\lcMatchKept{29}
\def\lcMatchTok{282}
\def\lcMatchKM{+0.31}

\def\lcMatchLM{-0.01}

\def\lcSumM{+0.03}
\def\lcSumLo{-0.06}
\def\lcSumHi{+0.12}
\def\lcSumKept{18}
\def\lcSumTok{282}
\def\lcSumKM{+0.25}

\def\lcSumLM{-0.02}

\def\lcStrM{-0.04}
\def\lcStrLo{-0.13}
\def\lcStrHi{+0.04}
\def\lcStrKept{13}
\def\lcStrTok{216}
\def\lcStrKM{+0.00}

\def\lcStrKN{17}
\def\lcStrLM{-0.05}

\def\lcFirstM{-0.02}
\def\lcFirstLo{-0.08}
\def\lcFirstHi{+0.04}
\def\lcFirstKept{1}
\def\lcFirstTok{282}
\def\lcFirstLM{-0.02}

\def\lcLostHelpSum{6}
\def\lcLostNSum{111}

\def\lcMems{57}
\def\lcPoolKeptM{+0.32}
\def\lcPoolKeptLo{+0.17}
\def\lcPoolKeptHi{+0.49}

\def\lcPoolLostM{-0.02}
\def\lcPoolLostLo{-0.08}
\def\lcPoolLostHi{+0.03}

\def\lcBearCells{135}
\def\lcBearTasks{55}
\def\lcNullRows{6}
\def\lcRawLenR{-0.13}
\def\lcPosChiP{0.09}
\def\lcPosStd{+0.39}
\def\lcPosUnadj{+0.39}

\def\lcSelSumKept{+0.38}
\def\lcSelSumLost{+0.34}
\def\lcSelMatchKept{+0.33}
\def\lcSelMatchLost{+0.35}
\def\lcSelStrKept{+0.18}
\def\lcSelStrLost{+0.37}

\def\lfScrCacheHits{0} 
\def\lfScrRealCalls{5{,}198} 
\def\lfHurtClBearLo{0.0} 
\def\lfHurtClBearHi{31.2} 
\def\lfHelpClBearLo{31.4} 
\def\lfHelpClBearHi{50.4} 
\def\lfHurtClCtrlLo{10.3} 
\def\lfHurtClCtrlHi{50.0} 
\def\lfHelpClCtrlLo{3.7} 
\def\lfHelpClCtrlHi{12.4} 
\def\lfHurtClRandLo{9.3} 
\def\lfHurtClRandHi{19.9} 
\def\lfHelpClRandLo{9.9} 
\def\lfHelpClRandHi{13.8} 
\def\lfTwoWayLo{+0.1675} 
\def\lfTwoWayHi{+0.4683} 
\def\lfSmCellsAll{285} 
\def\lfSmNullRows{20} 
\def\lfScrPosStd{+0.33} 
\def\lfCondFlipMems{35} 
\def\lfCondFlipMean{0.33} 
\def\lfCondFlipPZero{70} 
\def\lfPrLcCompared{27} 
\def\lfPrLcMismatch{3} 
\def\lfPrWfSumNew{-0.004} 
\def\lfPrWfSumOld{+0.004} 
\def\lfPrWfStrNew{-0.007} 
\def\lfPrWfStrOld{+0.011} 
\def\lfHeadD{+0.33} 
\def\lfHeadLo{+0.21} 
\def\lfHeadHi{+0.46} 
\def\lfHeadEpLo{+0.20} 
\def\lfHeadEpHi{+0.47} 
\def\lfHeadMems{60} 
\def\lfScrEpisodes{35} 
\def\lfScrTasks{218} 
\def\lfFillerK{4} 
\def\lfFillerLo{1{,}500} 
\def\lfFillerHi{5{,}000} 
\def\lfFillerTraj{50} 
\def\lfPoolEnt{1{,}024} 
\def\lfPoolWeb{865} 
\def\lfSizeCapDropped{11} 
\def\lfSizeCap{20{,}000} 
\def\lfMemWeb{40} 
\def\lfMemEnt{20} 
\def\lfLcCacheHits{14} 
\def\lfGfTaskLo{+0.05} 
\def\lfGfTaskHi{+0.70} 
\def\lfGfTaskN{44} 
\def\lfGfEpLo{+0.14} 
\def\lfGfEpHi{+0.67} 
\def\lfGfEpN{19} 
\def\lfGfCtlCells{42} 
\def\lfGfMemN{24} 
\def\lfGfMemD{+0.49} 
\def\lfGfJudge{gpt-oss-120b} 
\def\lfGfPicks{42} 
\def\lfGfGold{266} 
\def\lfGfBoth{10} 
\def\lfGfPrec{23.8} 
\def\lfGfRec{3.8} 
\def\lfDriftN{1{,}124} 
\def\lfDriftReward{82} 
\def\lfSweDemoted{37} 
\def\lfSweDemDjango{27} 
\def\lfSweDemSympy{10} 
\def\lfSweMatched{40} 
\def\lfJudgeObs{3{,}486} 
\def\lfJudgeUnits{60} 
\def\lfUnsHwN{22} 
\def\lfUnsOffN{22} 

\def\soXoneCalls{1{,}708}
\def\soXoneErr{2}
\def\soXoneCacheHits{0}
\def\soXoneCells{285}

\def\soXoneD{+0.50}
\def\soXoneLo{+0.37}
\def\soXoneHi{+0.63}
\def\soXoneMems{60}
\def\soXoneEpLo{+0.34}
\def\soXoneEpHi{+0.67}
\def\soXoneEps{35}
\def\soXoneTw{+0.40}
\def\soXoneTwLo{+0.26}
\def\soXoneTwHi{+0.55}
\def\soPosCells{284}
\def\soPosChg{28}
\def\soPosChgPct{10}
\def\soPosMov{81}
\def\soPosChgMv{28}
\def\soPosChgMvPct{35}
\def\soPosVarPct{11}
\def\soPosEdge{+0.43}
\def\soPosMid{+0.30}

\def\soStabSamePct{91}
\def\soStabSameK{0.78}
\def\soStabSameN{156}
\def\soStabDiffPct{87}
\def\soStabDiffK{0.65}
\def\soStabDiffN{697}

\def\soXtwoCells{277}
\def\soXtwoMems{57}
\def\soXtwoCollCells{63}
\def\soXtwoPad{8{,}065}
\def\soFRaw{+0.35}
\def\soFRawLo{+0.23}
\def\soFRawHi{+0.49}
\def\soFN{77}
\def\soFSum{+0.04}
\def\soFSumLo{-0.04}
\def\soFSumHi{+0.12}
\def\soFN{77}
\def\soFSumR{-0.31}
\def\soFSumRLo{-0.44}
\def\soFSumRHi{-0.18}
\def\soFStr{+0.01}
\def\soFStrLo{-0.04}
\def\soFStrHi{+0.07}
\def\soFN{77}
\def\soFStrR{-0.34}
\def\soFStrRLo{-0.47}
\def\soFStrRHi{-0.21}
\def\soFTr{+0.04}
\def\soFTrLo{-0.01}
\def\soFTrHi{+0.11}
\def\soFN{77}

\def\soFRawAll{+0.37}
\def\soFRawAllLo{+0.25}
\def\soFRawAllHi{+0.51}
\def\soFNAll{135}
\def\soFSumAll{+0.03}
\def\soFSumAllLo{-0.03}
\def\soFSumAllHi{+0.09}
\def\soFNAll{135}
\def\soFSumRAll{-0.34}
\def\soFSumRAllLo{-0.48}
\def\soFSumRAllHi{-0.21}
\def\soFStrAll{-0.01}
\def\soFStrAllLo{-0.07}
\def\soFStrAllHi{+0.03}
\def\soFNAll{135}
\def\soFStrRAll{-0.39}
\def\soFStrRAllLo{-0.52}
\def\soFStrRAllHi{-0.26}
\def\soFTrAll{+0.11}
\def\soFTrAllLo{+0.03}
\def\soFTrAllHi{+0.20}
\def\soFNAll{135}

\def\soScrZero{15}
\def\soScrOver{205}
\def\soScrDisc{75}
\def\soNoCapMin{18}
\def\soNoCapMax{91}
\def\soNoCapMems{1{,}889}
\def\soChanceTen{1}
\def\soChanceTenP{0.01}
\def\soRBmOne{0.03}

\def\soRBmTen{0.21}
\def\soRBmTenLo{0.14}
\def\soRBmTenHi{0.28}
\def\soRBmMrr{0.18}

\def\soRBmAuc{0.86}

\def\soRPreOne{0.11}

\def\soRPreTen{0.35}
\def\soRPreTenLo{0.26}
\def\soRPreTenHi{0.44}
\def\soRPreMrr{0.36}

\def\soRPreAuc{0.82}

\def\soRChkOne{0.14}

\def\soRChkTen{0.40}
\def\soRChkTenLo{0.31}
\def\soRChkTenHi{0.50}
\def\soRChkMrr{0.41}

\def\soRChkAuc{0.85}

\def\soRChkTenPct{40}

\def\rvDecCells{284}
\def\rvDecDf{105}
\def\rvIntShare{32}
\def\rvAddShare{55}
\def\rvEpsShare{13}
\def\rvIntF{8.5}
\def\rvIntP{<0.001}
\def\rvHarmMdePct{44}
\def\rvHarmRandPct{14.5}
\def\rvRandFireN{285}
\def\rvRandFireHelp{15.5}
\def\rvRandFireHelpN{220}
\def\rvRandFireHurt{10.8}

\def\rvRandSilentHelp{10.1}
\def\rvRandSilentHelpN{473}
\def\rvRandSilentHurt{16.7}
\def\rvGfUniverse{277{,}594}
\def\rvGfCand{841}
\def\rvGfYes{42}
\def\rvGfTasks{23}
\def\rvGfMems{28}
\def\rvSciRaw{-0.01}
\def\rvSciRawLo{-0.05}
\def\rvSciRawHi{+0.02}
\def\rvSciRawTied{122}
\def\rvSciRawN{127}

\def\rvSweRaw{-0.05}
\def\rvSweRawLo{-0.17}
\def\rvSweRawHi{+0.07}
\def\rvMatchExact{57}
\def\rvMatchNorm{82}
\def\rvFailExactPct{60}
\def\rvFailNormPct{50}
\def\rvRetried{1}
\def\rvScrRows{2{,}599}
\def\rvCapFourTasks{47}
\def\rvCapFourMems{64}

\def\rvCapEightTasks{75}
\def\rvCapEightMems{165}

\def\rvCapSixteenTasks{88}
\def\rvCapSixteenMems{241}

\def\rvCapThirtyTwoTasks{101}
\def\rvCapThirtyTwoMems{355}

\def\rvCapShareLo{64}
\def\rvCapShareHi{72}
\def\rvSelSingle{25}
\def\rvSelMulti{35}
\def\rvPopMulti{46}
\def\rvPopAll{165}
\def\rvHeadSingle{+0.36}
\def\rvHeadSingleLo{+0.12}
\def\rvHeadSingleHi{+0.60}
\def\rvHeadMulti{+0.32}
\def\rvHeadMultiLo{+0.19}
\def\rvHeadMultiHi{+0.45}
\def\rvHeadW{+0.35}
\def\rvHeadWLo{+0.17}
\def\rvHeadWHi{+0.53}
\def\rvPopSinglePct{72}

\definecolor{tealdeep}{HTML}{178A86}
\definecolor{teallight}{HTML}{3DB5AC}
\definecolor{coral}{HTML}{E2553F}

\makeatletter
\newif\if@qmtitle
\newcommand{\qmark}{?\global\@qmtitletrue}
\def\showarticletitle#1{\global\@qmtitlefalse#1\if@qmtitle\expandafter\@gobbleperiod\fi}
\def\@gobbleperiod{\@ifnextchar.{\@gobble}{}}
\makeatother

\definecolor{ink}{HTML}{3A3A4A}

\newcommand{\supp}[1]{Supplement~\ref*{S-#1}}

\setcopyright{none}
\acmConference[Preprint]{Preprint}{October 2026}{Online}{}
\copyrightyear{2026}
\acmYear{2026}
\acmDOI{}
\acmISBN{}

\title{What Is This Memory Useful For? Measuring Task-Conditioned
Utility in Long-Term Agent Memory}

\author{Chih-Hsuan Yang}
\authornote{These authors are co-first authors and contributed equally to this work.}
\authornote{Corresponding author: Chih-Hsuan Yang (bellayang@anl.gov).}
\affiliation{%
  \institution{Argonne National Laboratory}
  \city{Lemont}
  \state{Illinois}
  \country{USA}
}
\email{bellayang@anl.gov}

\author{Anjir Ahmed Chowdhury}
\authornotemark[1]
\affiliation{%
  \institution{University of Houston}
  \city{Houston}
  \state{Texas}
  \country{USA}
}

\author{Feng Yan}
\affiliation{%
  \institution{University of Houston}
  \city{Houston}
  \state{Texas}
  \country{USA}
}
\email{fyan5@uh.edu}

\author{Rajeev Thakur}
\affiliation{%
  \institution{Argonne National Laboratory}
  \city{Lemont}
  \state{Illinois}
  \country{USA}
}
\email{thakur@anl.gov}

\begin{document}

\begin{abstract}
A stored experience may answer one future question and offer little help
with another. We measure this task-conditioned utility rather than assigning
each memory a single importance score. A task-blind writer stores a memory
without seeing the later task; a frozen reader then answers that task with
and without the memory, using identical filler. A cost ledger records the
resources used to form, store and expose the memory.

On single-call factual-recall tasks built from LongMemEval-V2 web-agent
memories, the same memory helps more when it contains information needed for
the answer. Its mean paired reward gain on these answer-bearing task pairs
exceeds that on matched-control pairs by $\lfHeadD$ (95\% interval
$\lfHeadLo$ to $\lfHeadHi$). The tested task-blind summaries and structured records yield smaller gains
than Raw, including at fixed total context; a budget sweep is consistent
with retention of answer-bearing content contributing to the difference.
The bearing--control contrast remains positive in a new session with
randomized memory position and under a second reader, although individual
memory--task labels change. Generic retrieval performs above
chance but misses most screen-defined answer-bearing memories. The factual-recall gains do not extend to the tested software-engineering
and scientific-coding benchmarks, where even raw memories have no measurable
effect. We provide a public, versioned
\href{https://huggingface.co/datasets/AgentsSci/AAMAS2027-ConditionalMemory}{Conditional Memory Utility Dataset}
of paired measurements and costs. The practical takeaway is to evaluate memory compression by the task
utility it preserves alongside its costs. Task-conditioned measurements
make this comparison possible; their stability must also be checked.
\end{abstract}



\begin{teaserfigure}
\noindent\makebox[\textwidth][c]{\small
\href{https://chihhsuan-yang.github.io/AAMAS2027-ConditionalMemory/}{Project Website}\quad$\vert$\quad
\href{https://github.com/ChihHsuan-Yang/AAMAS2027-ConditionalMemory}{Code}\quad$\vert$\quad
\href{https://huggingface.co/datasets/AgentsSci/AAMAS2027-ConditionalMemory}{Dataset}
}
\Description{Links to the project website, code, and dataset.}
\end{teaserfigure}

\maketitle

\section{Introduction}

A long-horizon agent has to decide which past experiences to retain, which to
retrieve for a task, and which to expose to the model that answers it. For
retention, the relevant quantity is a memory's expected effect over future
tasks; for exposure, it is the effect on the task at hand. Both depend on how a
memory's effects vary across tasks: whether it helps many tasks or few, and
whether a memory that helps one task can hurt another. Memory systems already score items~\cite{multifactor,memoryrone,umem}, and prior
work observes that value depends on the task~\cite{ember,evomemeval,meclear}. We provide a measurement protocol that jointly holds the filler context
fixed, keeps the writer blind to the later task, and records the memory's
cost beside its effect. We apply it to web-agent memory on
LongMemEval-V2~\cite{lmev2}. The central finding is that usefulness depends
on the memory--task pair and on whether the representation preserves the
content needed for that task.

Consider a hypothetical past page visit that shows a project's budget. A
task-blind writer stores that visit as a memory without seeing any future
question. A later question asks for the budget, and the same reader answers it
twice, with the same filler, once with and once without the stored memory; the
reward difference is that memory's effect on that question. Repeating the
comparison for other questions and representations yields the memory's
task-conditioned profile.

The controlled comparisons show that a memory gains more on tasks for
which it contains answer-bearing information than on matched controls.
The tested summaries and structured records gain less than Raw, including
when total context is fixed. These findings make preserved task utility a concrete evaluation target
for memory compression, alongside its costs: a shorter memory is useful
only if it still helps the reader answer the relevant question. They also expose two
practical challenges: generic retrieval misses most screen-defined
answer-bearing memories, and individual memory--task labels can change
across measurement conditions. The aggregate bearing--control contrast
remains positive across the tested sessions, positions and readers;
that stability does not make each individual label stable.

\paragraph{Scope.}
Our empirical evidence comes from single-call factual-recall evaluation: the
memories are page visits from web-agent trajectories, the future evaluation
tasks are questions with deterministically scored answers, and a reader answers
each question in one call without acting in an environment. We make no claim
about procedural memory, acting agents, or a trained retention policy.

\paragraph{Contributions.}
\textbf{(1) A conditional memory utility protocol} (Section~\ref{sec:method}):
paired evaluation with identical context filler, a memory writer that never
observes the future task, and a cost ledger that records formation, storage,
retrieval and exposure costs beside each measurement.
\textbf{(2) Bounded findings for factual recall} (Sections~\ref{sec:screened},
\ref{sec:similarity}, \ref{sec:repr}; Figures~\ref{fig:robust}
and~\ref{fig:mech}). Within the same memory, answer-bearing task pairs gain
$\lfHeadD$ $[\lfHeadLo, \lfHeadHi]$ more reward than matched-control pairs. The
evaluated task-blind compressed representations show smaller gains than Raw at
their own length and at a fixed total context, consistent with reduced
retention of answer-bearing text, and generic retrieval recovers only part of
the screen-defined answer-bearing set ($\soRChkTenPct\%$ in its top ten, against
about $\soChanceTen\%$ by chance).
\textbf{(3) Measurement-reliability findings} (Sections~\ref{sec:metrics},
\ref{sec:readertransfer}, \ref{sec:r0}, \ref{sec:limitations}). The headline
contrast keeps its direction across a new session with randomized position
($\soXoneD$) and a second reader ($\rd{\rgBmcRawGem}$), while its magnitude and
per-pair labels change; sign-flip and harm summaries are confounded by pairs per
memory and baseline correctness, and we show how to condition both away.
\textbf{(4) A dataset} (Section~\ref{sec:release}) of per-pair measurements with
costs, for reproducing the contrasts, evaluating retrieval against
screen-defined labels, and studying variability.

\section{Related Work}
\label{sec:related}
We organize prior work by the granularity of memory-utility measurement and
the evaluation controls it provides. Several systems are designed around the
idea that a memory's value varies across tasks. A design rationale alone does
not supply what a retention policy needs:
how large the per-task effect is against execution noise, how often it
occurs, and what the memory costs.

\paragraph{Measured, then averaged.}
UMEM scores a memory by a marginal-utility reward over a cosine neighborhood
of $N{=}3$ future queries under GRPO~\cite{umem}. The counterfactual is real,
but it is reduced to its mean, and no dispersion or negative-utility
statistic is reported. Their Lemma~C.1 argues that the neighborhood captures
the queries that retrieve the same memory; it does not report whether the
dispersion inside that population is structured, which is what we measure.

\paragraph{One memory, several future tasks.}
EMBER's MultiQ split tests whether one frozen memory serves 100 bundles of
five questions each~\cite{ember}. EvoMemEval releases per-cell labels from
real execution with and without the memory on held-out tasks~\cite{evomemeval}; the labels
are categorical, so they carry no effect size, and each cell runs once, so no
flip separates from execution variance. Its released balance (exactly
100/100/100 by seeded down-sampling) is itself a design choice, so a rate
computed on it describes the sample rather than a population. MeClear computes leave-one-out and
Shapley attributions from re-execution, but over the memories available to a
single query~\cite{meclear}. A multi-factor value model notes that task
utility is among the factors its scorer leaves out~\cite{multifactor}, and Memory-R1 trains a memory manager with one
task per memory~\cite{memoryrone}.

\paragraph{Constraints on our design.}
Memory-R2 argues that once rollouts write, update or delete different
memories, trajectory-level comparisons become unfair~\cite{memoryrtwo}; our
frozen memory set answers this in Section~\ref{sec:method}. Selective
retention on clean benchmarks finds retention policies indistinguishable
within Wilson 95\% intervals~\cite{traceretain}, which is expected if no
memory is harmful. In EvoMemEval's released sample, pooled over its three datasets, the
\texttt{harmful} label covers 22.0\% of 537 items; we did not validate these
labels. A learned gate over measurements like ours could build on
influence-guided context selection~\cite{influence} and pre-generation
decoding of tool necessity~\cite{when2tool}; we propose none. Leave-one-out
re-execution is shared with MeClear, UMEM and influence-guided
selection; context attribution by ablation is developed further in
ContextCite~\cite{contextcite}, and per-item valuation in data
Shapley~\cite{datashapley}. What we add is the measurement design around it: profiles across tasks under
controlled context, formation-cost provenance, and sensitivity checks. Our instability findings also have known causes. Long-context readers are
sensitive to where relevant text sits~\cite{lostmiddle}, and temperature-0
inference is not reproducible when serving kernels vary with batch
composition~\cite{nondeterminism}. What is new is measuring their size on the quantity
a memory evaluation reports. On
representation, LongMemEval's original study already varied the granularity
at which history is stored and retrieved~\cite{lmev1}; our length control
asks a narrower question, whether a representation's effect comes from what it
says or from how many tokens it keeps.

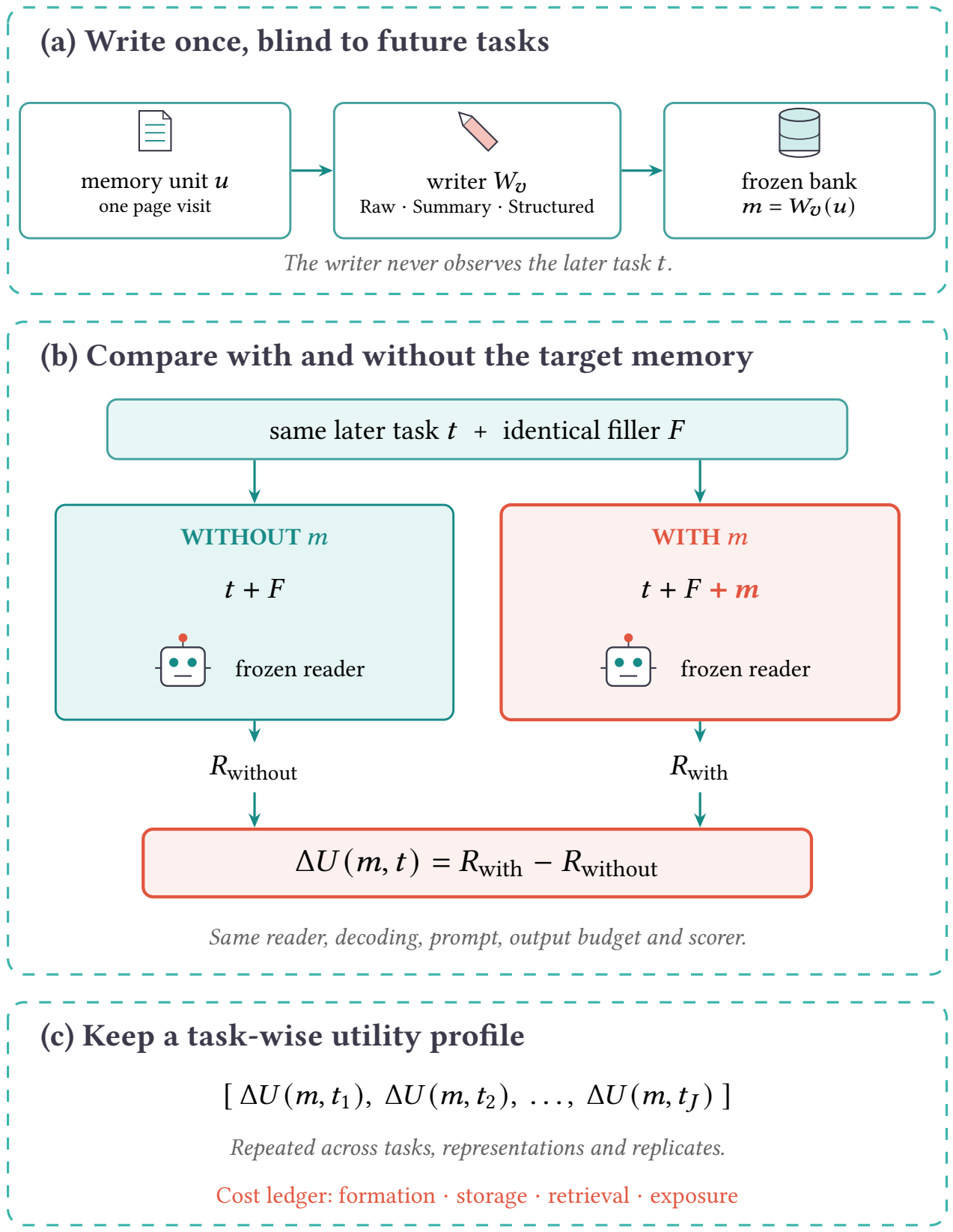
\begin{figure}[t]
\centering
\begin{tikzpicture}[x=1pt,y=1pt,
  panel/.style={draw=teallight, dashed, rounded corners=4pt, line width=0.6pt},
  ptitle/.style={anchor=west, font=\small\bfseries, text=ink},
  box/.style={draw=tealdeep!80, rounded corners=2.5pt, line width=0.55pt, fill=white, align=center},
  note/.style={font=\scriptsize\itshape, text=black!62},
  flow/.style={-{Stealth[length=3.6pt,width=3pt]}, draw=tealdeep, line width=0.6pt},
  pageicon/.pic={\draw[ink, line width=0.4pt, fill=white] (-4,-5) -- (-4,5) -- (2,5) -- (4,3) -- (4,-5) -- cycle;
                 \draw[ink, line width=0.3pt] (2,5) -- (2,3) -- (4,3);
                 \foreach \yy in {1.5,-0.5,-2.5} \draw[tealdeep, line width=0.35pt] (-2.5,\yy) -- (2.5,\yy);},
  bankicon/.pic={\draw[ink, line width=0.4pt, fill=teallight!25] (-5,-5) rectangle (5,4);
                 \draw[ink, line width=0.4pt, fill=teallight!25] (0,4) ellipse (5 and 1.6);
                 \draw[ink, line width=0.3pt] (-5,-0.5) arc (180:360:5 and 1.6);
                 \draw[ink, line width=0.4pt] (-5,-5) arc (180:360:5 and 1.6);},
  robot/.pic={\draw[ink, line width=0.35pt] (0,5.2) -- (0,7.2); \fill[coral] (0,7.6) circle (1.1);
              \draw[ink, line width=0.45pt, fill=white, rounded corners=1.6pt] (-5.5,-4.5) rectangle (5.5,5.2);
              \fill[tealdeep] (-2.3,1.4) circle (1.15); \fill[tealdeep] (2.3,1.4) circle (1.15);
              \draw[ink, line width=0.4pt] (-2.2,-2) -- (2.2,-2);
              \draw[ink, line width=0.4pt] (-5.5,0) -- (-7,0); \draw[ink, line width=0.4pt] (5.5,0) -- (7,0);},
  pencil/.pic={\draw[ink, line width=0.4pt, fill=coral!35, rotate=45] (-1.6,-5.5) rectangle (1.6,3.5);
               \draw[ink, line width=0.4pt, fill=white, rotate=45] (-1.6,3.5) -- (0,6.5) -- (1.6,3.5);}]
\draw[panel] (0,0) rectangle (236,-72);
\node[ptitle] at (5,-9) {(a) Write once, blind to future tasks};
\node[box, minimum width=68pt, minimum height=34pt] (u) at (37,-41) {};
\pic at (37,-31) {pageicon};
\node[font=\scriptsize, align=center] at (37,-47) {memory unit $u$\\[-1pt]{\tiny one page visit}};
\node[box, minimum width=72pt, minimum height=34pt] (w) at (118,-41) {};
\pic at (118,-31) {pencil};
\node[font=\scriptsize, align=center] at (118,-47) {writer $W_v$\\[-1pt]{\tiny Raw $\cdot$ Summary $\cdot$ Structured}};
\node[box, minimum width=68pt, minimum height=34pt] (b) at (199,-41) {};
\pic at (199,-31) {bankicon};
\node[font=\scriptsize, align=center] at (199,-47) {frozen bank\\[-1pt]{\tiny $m = W_v(u)$}};
\draw[flow] (u.east) -- (w.west); \draw[flow] (w.east) -- (b.west);
\node[note] at (118,-64) {The writer never observes the later task $t$.};
\draw[panel] (0,-79) rectangle (236,-243);
\node[ptitle] at (5,-88) {(b) Compare with and without the target memory};
\node[box, minimum width=186pt, minimum height=15pt, font=\footnotesize, fill=teallight!12] (top) at (118,-106)
  {same later task $t$ \;+\; identical filler $F$};
\node[box, draw=tealdeep, fill=teallight!14, minimum width=100pt, minimum height=54pt] (wo) at (62,-152) {};
\node[box, draw=coral, line width=0.8pt, fill=coral!8, minimum width=100pt, minimum height=54pt] (wi) at (174,-152) {};
\node[font=\scriptsize\bfseries, text=tealdeep] at (62,-133) {WITHOUT $m$};
\node[font=\scriptsize\bfseries, text=coral] at (174,-133) {WITH $m$};
\node[font=\footnotesize] at (62,-146) {$t + F$};
\node[font=\footnotesize] at (174,-146) {$t + F$ \textcolor{coral}{$\boldsymbol{+\,m}$}};
\pic at (44,-166) {robot}; \node[font=\scriptsize, anchor=west] at (54,-166) {frozen reader};
\pic at (156,-166) {robot}; \node[font=\scriptsize, anchor=west] at (166,-166) {frozen reader};
\draw[flow] (top.south -| wo.north) -- (wo.north);
\draw[flow] (top.south -| wi.north) -- (wi.north);
\node[font=\footnotesize] (rwo) at (62,-191) {$R_{\text{without}}$};
\node[font=\footnotesize] (rwi) at (174,-191) {$R_{\text{with}}$};
\draw[flow] (wo.south) -- (rwo.north); \draw[flow] (wi.south) -- (rwi.north);
\node[box, draw=coral, line width=0.8pt, fill=coral!8, minimum width=168pt, minimum height=17pt, font=\small] (du)
  at (118,-215) {$\Delta U(m,t) = R_{\text{with}} - R_{\text{without}}$};
\draw[flow] (rwo.south) -- (rwo.south |- du.north);
\draw[flow] (rwi.south) -- (rwi.south |- du.north);
\node[note] at (118,-234) {Same reader, decoding, prompt, output budget and scorer.};
\draw[panel] (0,-250) rectangle (236,-306);
\node[ptitle] at (5,-259) {(c) Keep a task-wise utility profile};
\node[font=\footnotesize] at (118,-274) {$[\,\Delta U(m,t_1),\ \Delta U(m,t_2),\ \dots,\ \Delta U(m,t_J)\,]$};
\node[note] at (118,-287) {Repeated across tasks, representations and replicates.};
\node[font=\scriptsize, text=coral] at (118,-299) {Cost ledger: formation $\cdot$ storage $\cdot$ retrieval $\cdot$ exposure};
\end{tikzpicture}
\caption{Conditional memory utility evaluation. \textbf{(a)}~Page-visit memory
units extracted from past episodes are stored as Raw, Summary or Structured
representations without exposing the writer to future evaluation tasks.
\textbf{(b)}~For a memory--task pair, a frozen reader is evaluated with identical
filler $F$, either with or without the target memory $m$; the paired reward
difference defines $\Delta U(m,t)$ under the recorded evaluation conditions.
\textbf{(c)}~Repeating this comparison across tasks, representations and
replicates yields a task-conditioned utility profile. Formation, storage,
retrieval and exposure costs are recorded alongside the measurements (the
cost ledger).}
\label{fig:pipeline}
\Description{A three-panel diagram. Panel a: a memory unit (one page visit) passes to a writer that produces Raw, Summary or Structured text, stored in a frozen bank; the writer never observes the later task. Panel b: the same later task and identical filler feed two arms, without and with the target memory, each answered by the same frozen reader; their rewards give Delta U equals R with minus R without. Panel c: repeating this gives a profile of Delta U over tasks, with a cost ledger of formation, storage, retrieval and exposure.}
\end{figure}

\section{Conditional Memory Evaluation}
\label{sec:method}

\subsection{The measured quantity}
Figure~\ref{fig:pipeline} shows the protocol end to end. We call the model that answers a future evaluation task the \emph{reader},
and reserve ``agent'' for the motivating setting. A \emph{memory unit} $u$ is one
page visit from a past episode; a memory $m$ is the Raw, Summary or Structured
text written from $u$. For a memory $m$ and a later, distinct future task $t$, the conditional utility of
$m$ for $t$ is the change in reward that adding $m$ causes, given the context
filler $F$, the position $p$ at which $m$ is inserted, and the seed $s$:
\begin{equation}
\label{eq:du}
  \Delta U(m, t \mid F, p, s) = R(t \mid F \cup_p m; s) - R(t \mid F; s).
\end{equation}
A positive value means the memory helped and a negative value that it hurt; a
zero difference indicates no measured change in reward, not identical answers
or absence of useful information. We call these labels \emph{helped},
\emph{unchanged} and \emph{hurt}; each holds under its evaluation condition. Both arms fix the reader model, serving
cluster, decoding, prompt scaffold, output-token budget, and scoring function
and version; they differ only in whether $m$ is present. We call them the
\emph{with-target-memory} and \emph{without-target-memory} arms (\emph{with $m$} and
\emph{without $m$}); both contain the filler. Reward stays continuous
wherever the benchmark gives partial credit; in the screened run it is
binary. The per-cell estimate is the mean over replicates. We write $\Delta
U(m,t)$ when $F$, $p$ and $s$ are fixed, and the conditioning matters: a single
inserted item can change the answer for reasons unrelated to its content. The unit of analysis is the
\emph{cell}, a memory--task pair $(m,t)$ under one evaluation condition; a memory
has no single utility here, and its result is its profile across the tasks it
faced. We compare at three levels: the with-$m$ versus without-$m$ reward
difference of each pair; within each memory, the mean difference on
answer-bearing task pairs minus that on matched-control task pairs (the
headline contrast); and representations written from the same memory unit.

\paragraph{Identical filler, frozen memories.}
Both arms place the same controlled filler in the reader's context, and the
\textsc{with} arm adds $m$. The filler is $\lfFillerK$ memory units of
$\lfFillerLo$--$\lfFillerHi$ tokens each, drawn from $\lfFillerTraj$
trajectories disjoint from every trajectory that hosts a bearing memory, so it
never contains another unit that bears on the task. The \textsc{without} arm
is therefore not an empty-context arm. On LongMemEval's first version~\cite{lmev1} the oracle context is entirely
evidence, so ablating ``the memory'' there would delete the whole context. The memory set is written once, frozen and content-hashed before any
pairing exists, and no arm writes, updates or deletes a memory. The arms
differ by exactly one item from that frozen set, so the divergent-state
hazard raised by Memory-R2~\cite{memoryrtwo} does not arise. The position of
the added memory is not neutral: moving only that position changed the answer
in 8 of 12 cells of a pilot, and $\soPosChgPct\%$ of labels at scale
(Section~\ref{sec:limitations}). In the main runs we hold position fixed across replicates, vary it
across cells, and record it on every row (\supp{sec:position}). Every
contrast we report is therefore conditional on the recorded positions;
position variance is not inside our intervals.

\paragraph{The writer never sees the future task.}
Memory formation is a function of the memory unit alone, $m = W(u)$ and
never $m = W(u,t)$. Without this invariant, $\Delta U > 0$ would only show
that a model can be told an answer and copy it forward. We enforce it
structurally rather than by convention: the writer takes a segment and
nothing else, emits a closed field set, and has a single interpolation point.
Each of these barriers carries a negative control that plants the defect and requires
the check to fire (\supp{sec:taskblind}).

\subsection{The answer-bearing screen is an instrument, not a retriever}
\label{sec:screen}
To decide which (memory, task) pairs are worth executing, we use an
\emph{answer-bearing screen}. For each pair it reads three inputs: the memory's text,
the task's \textbf{gold answer}, and the task's evaluation specification. It
never reads the question, a reward, or any model output, so the strata it
defines are not selected on the outcome. A pair is \emph{answer-bearing} (\textsc{bearing}) when the
benchmark's own scorer would mark the memory text correct against the gold
answer \emph{and} the test singles this memory out of its domain pool, firing
on at most eight of the 865--1{,}024 units available to that task.

Answer-bearing is a screening label based on answer containment, not a
measured positive-utility label: paired evaluation measures the effect, and
screen-selected pairs can still fail while pairs outside the screen can still
help. Our sparsity result is a statement about how answers are distributed
over memories. Because it consumes the gold answer, the screen
cannot be deployed, and we do not propose it as a retriever. Its role is to define, with the gold answer, the
answer-bearing pairs; these screen-defined labels, not exhaustive ground truth
for useful memories, are what we evaluate retrieval against (Section~\ref{sec:similarity}). Every rate we
report under the screen is conditional on a screened design.

A screen selected on the gold answer invites a second objection: perhaps the
effect is an artifact of selecting pairs with the answer in hand. We tested
this with a gold-free screen, pre-registered before execution: the same model
as the reader (\lfGfJudge), asked whether the memory contains the specific
information the question asks about, without the gold answer. On $\gfCells$
cells the gold screen \emph{never} selected, the gold-free picks move the
reader from $\gfWithoutPct\%$ correct without the memory to $\gfWithPct\%$ with
it, while $\lfGfCtlCells$ matched-control pairs move from
$\gfCtlWithoutPct\%$ to $\gfCtlWithPct\%$. The contrast is $\rd{\gfDiff}$, with a
task-clustered 95\% interval of $[\lfGfTaskLo, \lfGfTaskHi]$ over
$\lfGfTaskN$ tasks and an episode-clustered one of $[\lfGfEpLo, \lfGfEpHi]$
over $\lfGfEpN$ episodes. This answers the selection objection, not the
containment one; because the judge is the reader's own model, its picks may
favor pairs that model finds easy. Its picks overlap the gold screen little
(\supp{sec:goldfree}) yet carry an effect of similar size, so benefits also
occur outside the gold screen. A second, lexical gold-free screen did not
locate answer-bearing pairs (\supp{sec:goldfree}).

\subsection{Three representations of one experience}
All three representations derive from the \emph{same} memory unit, so a
downstream difference is attributable to how the segment is written rather
than to which content was selected. \textbf{Raw} is deterministic and uses no
model: the memory is the segment itself. \textbf{Summary} is one writer call
per segment at pinned decoding and prompt version. \textbf{Structured} is one
writer call that emits a validated schema (memory type, trigger, action,
outcome, lesson, scope). A malformed structure is recorded as a failure and
never repaired.

Formation cost belongs to the representation. Raw makes zero writer calls;
Summary and Structured make one each per segment. Omitting that cost flatters
the model-written representations, so we charge each its own, including
writes that failed after consuming a call.

\subsection{Noise, labels and cost}
\label{sec:metrics}
Noise is measured, not assumed. Replicates are separate requests
($\lfScrCacheHits$ of $\lfScrRealCalls$ calls served from cache), and within a
session they agree on all 1{,}157 primary cells, so the execution floor is zero
and a cell's label (helped, unchanged or hurt) is the sign of $\Delta U$; the pre-registered floors and
label rules are in \supp{sec:floors}. Truncated completions are re-issued, not
scored. Intervals are over memories, tasks or episodes, never replicates.

A memory shows a \textbf{sign flip} when it helps one task and hurts another. A flip rate is descriptive, not evidence of
memory\,$\times$\,task interaction: under a purely additive model
$\Delta U = \mu + a_m + b_t$, simulated over 20 trials of 200 memories
$\times$ 8 tasks, flip rates range from 53.5\% to 92.0\% with no interaction
present (\supp{sec:floors}). A flip rate also grows with the number of cells
per memory, so it can only be compared across groups at matched cell counts.

\paragraph{The perturbation floor.}
The noise that matters most here is not seed variation but the reader's
response to inserting \emph{any} item: position alone changes
$\soPosChgPct\%$ of per-cell labels (Section~\ref{sec:limitations}). We ran no placebo arm of length-matched irrelevant content,
so per-cell labels include such perturbations. The random-task stratum bounds their size: a memory paired with a randomly
selected task helps in $\raHelpRateRand\%$ of the pairs where the reader was
wrong without it and hurts in $\raHurtRateRand\%$ of those where it was right.
These rates are an upper bound on pure perturbation: the memories are real
same-domain pages, and where they contain the answer string they help more
often ($\rvRandFireHelp\%$ against $\rvRandSilentHelp\%$). Irrelevant
context can distract a reader~\cite{distracted} or even help it~\cite{powernoise}. We therefore report effects as paired contrasts between strata, which reduce
sensitivity to content-independent perturbation when its mean effect is
comparable across the contrasted strata, and never as counts of labeled cells;
without a placebo arm we do not claim to isolate every such effect.

\paragraph{Pre-registered and exploratory analyses.}
The screened design, both gold-free screens, the length control, the
second-reader study, the writer\,$\times$\,reader factorial, the reversal rate
and the variance decomposition were pre-registered; \supp{sec:prereg} lists
their anchors and every deviation. All other analyses, including the three
follow-up runs, are exploratory, and their intervals are not adjusted for
multiplicity.

Every measurement is paired with its cost in seven families, from storage to
formation; utility per exposed token is reported beside utility per token
charged, including the writer's (\supp{sec:reuse}).

\section{Benchmarks and Setup}
\label{sec:setup}

\paragraph{LongMemEval-V2 and the page-visit memory unit.}
Our primary benchmark is the \texttt{small} tier of
LongMemEval-V2~\cite{lmev2}, where each domain has exactly one haystack, so
every memory is available to every question in its domain. A whole
trajectory cannot serve as a memory: its median length is 103{,}886 tokens.
We segment at the \emph{page visit}, the maximal run of consecutive states
that share one URL, a boundary drawn by the past agent's own navigation. This
yields \textbf{1{,}889 memory units} with a median of 7{,}261 tokens. Removing
one unit removes a median 3.33\% of its trajectory and 0.058\% of the domain
haystack, and segmentation preserves answerability for 278 of 295
deterministically scored questions (94.2\%).

\paragraph{Future tasks and the executed design.}
A future task is one question with its evaluation specification. Of 451
questions, 295 are deterministically scored; we use the benchmark's scorer
verbatim, and an LLM-judged task raises an error rather than scoring zero.
Seeded pairing within domain gives a grid of 15{,}112 memory--task pairs. We
execute a \emph{screened design} over 60 memories, a floor fixed before
running by a power study: $\lfMemWeb$ from the web domain and $\lfMemEnt$ from
the enterprise domain (unit pools of $\lfPoolWeb$ and $\lfPoolEnt$), drawn from
$\lfScrEpisodes$ source episodes. A content-blind size cap fixed before any
outcome removed $\lfSizeCapDropped$ candidate memories longer than
$\lfSizeCap$ tokens. The 60 memories are every memory bearing on two or more
tasks that passed the cap ($\rvSelMulti$) and $\rvSelSingle$ that bear on one,
so the design over-samples multi-task memories to give reversal a chance. Each memory is paired with three strata of
future evaluation tasks: \emph{answer-bearing task pairs} (\textsc{bearing});
\emph{matched-control task pairs}, tasks of the same family as a bearing task
(static environment, dynamic environment or procedure) on which the screen does
not fire for that memory; and the \emph{random-task stratum} (\textsc{random}),
unscreened tasks drawn at random, over $\lfScrTasks$ tasks in all.
All three strata contain the identical 60 memories, so contrasts are paired
within memory rather than against a different set of control memories. The design has $\spCells$ cells (Table~\ref{tab:coverage}).

\paragraph{Secondary benchmarks.}
We also ran the protocol on SciCode~\cite{scicode} (reusable helper routines,
$\sciPairs$ donor$\to$recipient pairs) and SWE-bench Verified~\cite{swebench}
($\sweMatched$ interpretable cells); there memories show no effect even in raw
form, so we report them in \supp{sec:crossbenchtable}.

\paragraph{Execution.}
In the primary arms, the reader and the memory writer are both
\texttt{gpt-oss-120b} on one serving cluster, with low reasoning
effort, temperature 0 and \texttt{max\_tokens} 512; similarity uses
SFR-Embedding-\allowbreak Mistral. Section~\ref{sec:readertransfer} changes only the
reader, to \texttt{gemma-4-31B-it}. An empty or truncated completion raises
and the cell is re-run, because scoring it as zero would push $\Delta U$
negative and manufacture harmful memories. The full configuration, caching
and saturation checks are in \supp{sec:config}.

\begin{table}[t]
\centering\small
\caption{Every run a reported number comes from. Counts are memory--task pairs
times the conditions each pair was evaluated under (representations, arms or
replicates, as named). A dataset row is one replicate of one pair under one
condition and records both arms. Excluded rows had no producible reward and are
dropped, never scored as zero.}
\label{tab:coverage}
\setlength{\tabcolsep}{2.5pt}
\begin{tabular}{@{}llll@{}}
\toprule
run & memories & pairs $\times$ conditions & excluded \\
\midrule
\multicolumn{4}{@{}l}{\emph{LongMemEval-V2}} \\
screened design & \spMems & \spCells\ pairs & 0 rows \\
representations & \frMems & \frCells\ $\times$ 3 representations & 3 memories \\
length control & \lcMems & \frCells\ $\times$ 8 arms & \lcNullRows\ rows \\
second reader & \spMems & \rgSmCells\ of \lfSmCellsAll\ pairs & \lfSmNullRows\ rows \\
gold-free screen & -- & \gfCells\ $+$ \lfGfCtlCells\ pairs & -- \\
second session & \soXoneMems & \soXoneCells\ $\times$ 3 replicates & \soXoneErr\ rows \\
fixed context & \soXtwoMems & \soXtwoCells\ $\times$ 5 arms & \soXtwoCollCells\ pairs \\
retrieval & -- & \soScrDisc\ tasks (screen labels) & -- \\
\midrule
SciCode & \sciMems & \covScDone\ of \covScElig\ pair--repr. & \covScExcl\ pair--repr. \\
SWE-bench & -- & \covSwDone\ of \covSwElig\ pair--repr. & \covSwExcl\ pair--repr. \\
\bottomrule
\end{tabular}
\end{table}

\section{Results}
\label{sec:results}

\subsection{Specific answers are sparse over memories}
\label{sec:scarcity}
\begin{figure}[t]
\centering
\includegraphics[width=\columnwidth]{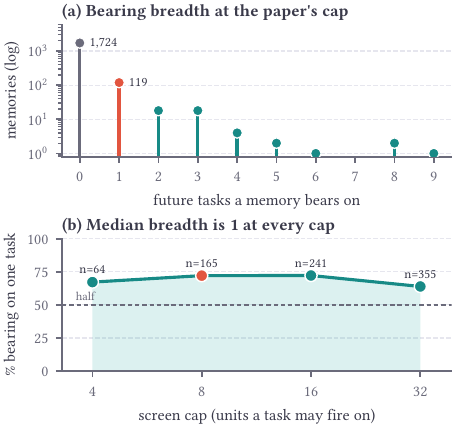}
\caption{Answer bearing is sparse at every screen cap (exploratory).
\textbf{(a)}~Number of memory units (of $\bearUnits$) that bear on each count of
screened future tasks, at the paper's cap of eight units per task (log axis).
\textbf{(b)}~Share of bearing memories that bear on exactly one task when the
cap is 4, 8, 16 or 32 units; $n$ is the number of bearing memories. The median
breadth is one at every cap.}
\label{fig:sparse}
\Description{Two panels. Top: a log-scale histogram of how many tasks each memory bears on, with 1,724 memories at zero and 119 at one, falling quickly after. Bottom: at screen caps of 4, 8, 16 and 32, between 64 and 72 percent of bearing memories bear on exactly one task.}
\end{figure}
Screening all 1{,}889 memory units against the 295 deterministically scored
questions yields 75 discriminative tasks: questions for which the screen fires
on at least one and at most eight units of the domain pool. Over those tasks,
165 memories bear on at least one, across 266 memory--task edges, and the
distribution is concentrated at one (Figure~\ref{fig:sparse}a): 119 memories
bear on exactly one task, 18 on two and 18 on three, with a tail reaching
nine. \emph{Among the 165 memory units that bear on at least one of the 75 qualifying
tasks, the median bears on one}; the other 1{,}724 units bear on none, so this
is not a median over all 1{,}889 units. Only 46 memories bear on two or more
tasks, and they come from 23 source trajectories, so they are not 46
independent draws; we report episode-clustered intervals wherever this
matters.

These are screen statistics, not re-execution results. The screen's cap of
eight units is what makes it specific: of the 220 deterministic questions it
excludes, $\soScrZero$ fire on no unit and $\soScrOver$ on more than eight,
often because short answers, such as a single number or letter, occur on many
pages. Without the cap, string containment alone fires for every one of
the $\soNoCapMems$ memories, on $\soNoCapMin$ to $\soNoCapMax$ tasks each. The pattern persists across the tested caps: at 4, 8, 16 and 32 units,
the median answer-bearing memory bears on exactly one task, and
$\rvCapShareLo$--$\rvCapShareHi\%$ bear on exactly one
(Figure~\ref{fig:sparse}b). The
sparsity we report therefore concerns \emph{specific} answers in this
benchmark, with these memory units and tasks. The screen does not identify
all useful pairs: cells outside \textsc{bearing} contribute more helped cells ($\raNonBearHelped$) than
bearing cells do (50), most at the perturbation floor
(Section~\ref{sec:metrics}), and the gold-free screen's picks lift accuracy from
$\gfWithoutPct\%$ to $\gfWithPct\%$ on cells the gold screen never selected
(Section~\ref{sec:screen}). A gold-free definition does not make bearing
denser, though: run over all $\rvGfUniverse$ (task, unit) pairs under the same
cap, the gold-free screen confirms only $\rvGfYes$.

Sparsity makes indiscriminate pairing uninformative. When each memory is paired
with same-domain questions chosen for question-type spread, without requiring
that it bear on them, the reader fails both with and without the memory in
376 of 479 and 402 of 480 cells in two prompt arms, about 80\%. Such a cell
yields $\Delta U = 0$ whether or not the memory is useful, so a null there is
an absence of evidence, not evidence of absence.

\subsection{Screening makes the effect measurable}
\label{sec:screened}
\begin{figure}[t]
\centering
\includegraphics[width=\columnwidth]{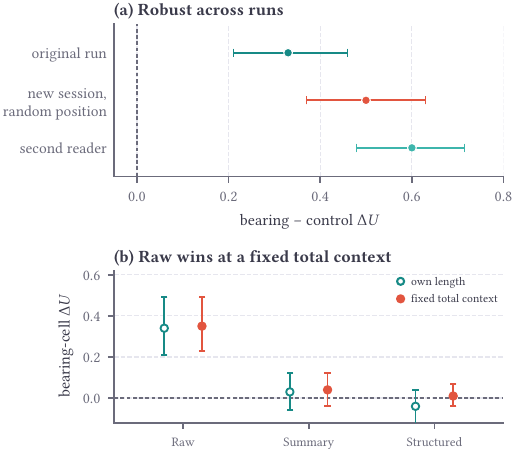}
\caption{The bearing--control contrast remains positive across tested
conditions (LongMemEval-V2). \textbf{(a)}~Bearing minus
control, paired within memory, in the original run, in a new session with the
memory's position redrawn per replicate, and under a second reader (95\%
intervals over memories). \textbf{(b)}~Bearing-cell $\Delta U$ of each
representation at its own length ($n=\lcBearCells$) and with every arm padded to
the Raw arm's token count ($n=\soFN$, excluding cells exposed to a cache-key
defect); task-clustered intervals. Pre-registered: the original run and the
second reader in (a); exploratory: the new session in (a) and panel (b).}
\label{fig:robust}
\Description{Two panels. Top: three interval estimates of the bearing-minus-control effect, for the original run, a new session with random position, and a second reader, all well above zero. Bottom: Raw near 0.35 and Summary and Structured near zero, both at their own length and at a fixed total context.}
\end{figure}

\begin{table}[t]
\centering\small
\caption{Screened run on LongMemEval-V2: the same 60 memories paired with
three strata of tasks. Both-arms-fail is the fraction of cells in which the
reader fails with \emph{and} without the memory; such cells cannot show a
memory effect. The criterion for a material reduction (\textsc{bearing}'s
Wilson upper bound below the comparator's point estimate) was fixed in code
before these numbers existed.}
\label{tab:screened}
\begin{tabular}{@{}lrcr@{}}
\toprule
stratum & both-arms-fail & Wilson 95\% & mean $\Delta U$ \\
\midrule
\textsc{bearing}      & \textbf{0.53} & [0.44, \textbf{0.61}] & \textbf{\rd{\stratBearV}} \\
matched control       & 0.79 & [0.71, 0.85] & \rd{\stratCtrlV} \\
\textsc{random}       & 0.70 & [0.67, 0.73] & \rd{\stratRandV} \\
\bottomrule
\end{tabular}
\end{table}
The screened design reduces these structural zeros: cells that fail in both
arms fall to 0.53 for \textsc{bearing}, against 0.79 for the matched control and
0.70 for \textsc{random} (Table~\ref{tab:screened}). The memory helped in
$\spBearPos$ of $\spBearN$ \textsc{bearing} cells ($\spBearRate\%$), against
$\spCtrlRate\%$ for the control and $\spRandRate\%$ for \textsc{random}, and
$\spZeroPct\%$ of all $\spCells$ cells are exact zeros.

\textbf{The headline contrast.} For each memory, we compare its mean paired
reward gain on answer-bearing task pairs with its mean gain on matched-control
task pairs. Averaged over the $\lfHeadMems$ memories, this contrast is $\lfHeadD$ $[\lfHeadLo, \lfHeadHi]$
(episode-clustered $[\lfHeadEpLo, \lfHeadEpHi]$; two-way over memories and
tasks $[\rd{\lfTwoWayLo}, \rd{\lfTwoWayHi}]$; Figure~\ref{fig:robust}a).
It is similar for single-task ($\rvHeadSingle$) and multi-task
($\rvHeadMulti$) memories, and $\rvHeadW$ $[\rvHeadWLo, \rvHeadWHi]$ reweighted
to all $\rvPopAll$ bearing memories.
It differs from the bearing stratum's own mean $\Delta U$ in
Table~\ref{tab:screened}, which is not paired within memory.
Because the screen selects pairs whose memory contains the answer, this result
first validates the instrument: the protocol detects the effect it should.

\textbf{The contrast holds across session, position and reader.} Re-running all
$\soXoneCells$ bearing and control cells in a new session, three replicates each
with the memory's position redrawn ($\soXoneCalls$ calls, none from cache),
gives $\soXoneD$ $[\soXoneLo, \soXoneHi]$; under a second reader it is
$\rd{\rgBmcRawGem}$ (Section~\ref{sec:readertransfer}). The contrast keeps its
direction in every tested condition, while its magnitude moves with session and
reader, as Section~\ref{sec:limitations} anticipates.

Two checks address alternative explanations. First, the screen did not
select easier tasks. Because reward is binary, Table~\ref{tab:screened} and the
cell counts fix each stratum's accuracy without the memory: $\raAccZeroBear\%$
for \textsc{bearing}, $\raAccZeroCtrl\%$ for the control and
$\raAccZeroRand\%$ for \textsc{random}. Bearing tasks are the hardest without
the memory and the easiest with it ($\raAccOneBear\%$). The
$\raHurtOppBear$ bearing cells answered correctly without the memory
($\raBothCorrectBear$ of them also with it) show that the answer is sometimes
recoverable from the filler or from the reader's priors. Second, the reader can retrieve planted content: a unique code planted inside the target
memory, in the full $\sim$27k-token context, is recovered in 22 of 22 cells
and in 0 of 17 when nothing is planted (\supp{sec:needle}). This makes a simple
inability to access the long context less likely as the cause of low accuracy
on unscreened pairs.

\textbf{With the answer in context, the reader still fails on
$\raFailWithBearPct\%$ of bearing cells.} The planted-code test makes a simple
inability to access the context less likely, while failures to localize or use
natural evidence remain possible. These failures occur across question families (\supp{sec:lengthcontrol}).
Neither scorer leniency nor output truncation explains this pattern: every bearing pair matches the gold answer exactly ($\rvMatchExact$) or after
normalization ($\rvMatchNorm$), none by loose phrase containment, and exact
matches fail more often ($\rvFailExactPct\%$ against $\rvFailNormPct\%$).
Truncated outputs are re-run, not scored ($\rvRetried$ of $\rvScrRows$ rows).

\subsection{Retrievers find answer-bearing memories, but miss most}
\label{sec:similarity}
\begin{figure}[t]
\centering
\includegraphics[width=\columnwidth]{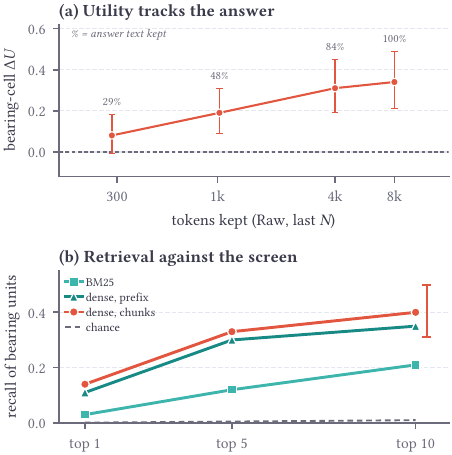}
\caption{What carries utility (LongMemEval-V2). \textbf{(a)}~Bearing-cell
$\Delta U$ as Raw is cut to its last $N$ tokens ($\lcBearCells$ cells,
task-clustered intervals); labels give the share of cells whose text still
contains the answer. \textbf{(b)}~Share of a task's bearing units that a
retriever ranks in its top 1, 5 and 10 of the domain pool ($\soScrDisc$ tasks;
task-clustered interval shown at top 10); the dashed line is chance. Both panels are
exploratory; the matched-length contrast in (a) was pre-registered.}
\label{fig:mech}
\Description{Two panels. Top: bearing-cell utility rising with the number of Raw tokens kept, with the share of cells keeping the answer rising from 29 to 100 percent. Bottom: recall at ten for BM25, prefix and chunked dense retrieval, from 0.21 to 0.40, far above a near-zero chance line.}
\end{figure}
The screen defines which units of a domain pool are answer-bearing for each
of the $\soScrDisc$ tasks, so these screen-defined labels can score retrievers
that never see the answer (Figure~\ref{fig:mech}b). Recall@10 measures recovery
of these units, not of every helpful memory or the accuracy of a complete
retrieval system. Ranking the whole pool of $\lfPoolWeb$ or $\lfPoolEnt$ units by
the question, dense retrieval with SFR-Embedding-Mistral ranks, on average,
$\soRChkTen$ $[\soRChkTenLo, \soRChkTenHi]$ of a task's bearing units in its top
ten (recall@10) when units are split into 512-token chunks, and $\soRPreTen$ when
only a 2{,}000-token prefix is embedded, against about $\soChanceTenP$ by chance.
The lexical ranker BM25 reaches $\soRBmTen$, with an area under the ROC curve
(AUROC) of $\soRBmAuc$. Chunked retrieval places the first
bearing unit at a mean reciprocal rank of $\soRChkMrr$, and chunking improves recall only modestly over prefix embedding in this
comparison. Retrieval thus locates bearing
memories far above chance yet misses most of them in its top ten, which is why
a deployable locator remains open. Similarity is a weaker guide to how
\emph{much} a memory helps: across executed cells, cosine similarity correlates
with $\Delta U$ at only $\rho=\rd{\simRhoAll}$ (\supp{sec:simdetail}).

\subsection{Evaluated compressed representations gain less than Raw}
\label{sec:repr}
We repeat the measurement on the same experiences and tasks, changing only how
each experience is written down. On the $\frCells$ cells evaluated under all
three representations, the paired \textsc{bearing}$-$control effect is
$\rd{\frRawDU}$ for Raw (95\% CI $[\rd{\frRawLo}, \rd{\frRawHi}]$).
For Summary it is $\rd{\frSumDU}$ ($[\rd{\frSumLo}, \rd{\frSumHi}]$), and
for Structured it is $\rd{\frStrDU}$ ($[\rd{\frStrLo}, \rd{\frStrHi}]$)
(Figure~\ref*{S-fig:overview}d in the supplement). Only Raw excludes zero. Per cell, Summary exposes
a mean $\frSumTokExp$ tokens, Structured $\frStrTokExp$ and Raw $\frRawTokExp$.
This Raw contrast comes from a different run and memory set than the screened
one (Table~\ref{tab:coverage}); the intervals overlap.

\textbf{The pre-registered length control.} A content-blind \emph{extractive
truncation} keeps the final $N$ tokens of the same segment, with $N$ set per cell to that
cell's Summary length. On the $\lcBearCells$ bearing cells (task-clustered over
$\lcBearTasks$ tasks), the extractive and Summary arms differ by $\rd{\lcPrimary}$
$[\rd{\lcPrimaryLo}, \rd{\lcPrimaryHi}]$, which does not establish equivalence,
and Raw beats the matched extractive arm by $\rd{\lcBudget}$
$[\rd{\lcBudgetLo}, \rd{\lcBudgetHi}]$. Keeping the first $N$ tokens instead of the last,
the contrast registered as decisive, retains the bearing content in 0.7\% of
cells against 28.9\%; last-$N$ exceeds first-$N$ by $\rd{\lcEnds}$
$[\rd{\lcEndsLo}, \rd{\lcEndsHi}]$.

\textbf{A budget sweep.} The same run also
truncated Raw to its last 4,096 and 1,024 tokens (Figure~\ref{fig:mech}a).
Bearing-pair gains fall with the budget, $\lcRawM$, $\lcFourM$, $\lcOneM$ and $\lcMatchM$
from full length down to matched length, and the share of cells whose text
still contains the bearing content falls with it: $\lcRawKept\%$,
$\lcFourKept\%$, $\lcOneKept\%$ and $\lcMatchKept\%$. Splitting every arm by
whether its text kept that content (Figure~\ref*{S-fig:lenctl}b), the arms look
alike: cells that kept it gain $\lcPoolKeptM$ $[\lcPoolKeptLo, \lcPoolKeptHi]$
pooled over the compressed arms, Summary included at $\lcSumKM$, and cells that
lost it gain $\lcPoolLostM$ $[\lcPoolLostLo, \lcPoolLostHi]$. The cells that
kept the content are not simply easier: on the very same cells, Raw gains
$\lcSelSumKept$ where Summary kept it and $\lcSelSumLost$ where Summary lost it.
Structured is the exception, at $\lcStrKM$ on $\lcStrKN$ cells, but those are
cells where even Raw gains only $\lcSelStrKept$. Within Raw, $\Delta U$ correlates weakly with memory length across bearing
cells ($r = \lcRawLenR$), which does not by itself establish the absence of a
length effect. Under the evaluated task-blind writing strategies, compressed
representations often omit answer-bearing text and show smaller reward gains
than Raw; because the sweep changes content and length together, it is
consistent with, rather than proof of, retention of answer-bearing content
driving the gains. These analyses are
exploratory (session caveats in \supp{sec:lengthcontrol}). The phrase matcher may miss paraphrased answers. However, Summary cells
classified as having lost the answer-bearing content help in only
$\lcLostHelpSum$ of $\lcLostNSum$. The result describes one task-blind writer
and prompt, whose structured schema targets procedural lessons; a writer told to
keep values and identifiers might keep more bearing content.

\textbf{Raw retains its advantage at a fixed total context.} In deployment the main
argument for compression is that a long raw memory displaces other context. We
re-ran the $\soXtwoCells$ representation cells with every arm of a cell given
the same token count: the compressed arms and the no-memory arm were padded with
further bearing-disjoint filler (median $\soXtwoPad$ tokens) up to the Raw arm's
total, the memory kept at its original position. On bearing cells Raw gains
$\soFRaw$ $[\soFRawLo, \soFRawHi]$, while Summary gains $\soFSum$ and Structured
$\soFStr$; against Raw on the same cells they lose $\soFSumR$
$[\soFSumRLo, \soFSumRHi]$ and $\soFStrR$ $[\soFStrRLo, \soFStrRHi]$. Including
cells exposed to a cache-key defect changes no conclusion (\supp{sec:sophia};
Figure~\ref{fig:robust}b). Costs depend on the accounting: amortizing the
writer's one-time cost over this design's $\frReuse$ evaluated uses per memory
removes compression's token advantage, and Summary overtakes Raw on point
estimates only above about $\raReuseStar$ uses per write, with a utility
interval spanning zero throughout (\supp{sec:reuse}).

\subsection{The contrast keeps its direction under a second reader}
\label{sec:readertransfer}
One concern is that the Raw effect is specific to one reader. We re-executed
the $\lfSmCellsAll$ Raw bearing and control cells with a second, unrelated reader
(gemma-4-31B-it) on the same memory text. The memory file, prompt, filler,
every rendered request and each target position were hash-asserted against
the first run, and the text re-verified byte-identical on $\rgRows$ rows.
Because this endpoint is not deterministic across sessions
(Section~\ref{sec:limitations}), we ran both readers in one session and quote
that measurement.

The aggregate contrast keeps its direction and is larger: over $\rgSmCells$ matched cells on
$\rgSmTasks$ tasks, the second reader's mean $\Delta U$ is higher by
$\rd{\rgSmDiff}$ $[\rd{\rgSmLo}, \rd{\rgSmHi}]$, and the within-memory
\textsc{bearing}$-$control contrast is $\rd{\rgBmcRawGem}$
(Figure~\ref{fig:robust}a). Part of the gain comes
from a weaker no-memory arm under the second reader ($\rd{\rgWithoutDelta}$)
rather than from a better memory.

\textbf{The aggregate contrast retains its direction across the two readers,
while per-pair labels show partial agreement.} Over the
$\rgSmCells$ cells, the two readers' three-way labels (helped, unchanged,
hurt) agree at Cohen's $\kappa = \raReaderKappa$. Of the $\raReaderMoving$
cells that move under either reader, both mark $\smAgree$ as helped
(Figure~\ref*{S-fig:overview}f). Too few cells are harmed ($\raHurtOss$ and
$\raHurtGem$) to tell whether harm labels reproduce. Whether a reader favors memories its own model
wrote is tested by a pre-registered writer\,$\times$\,reader factorial crossing
both models as writers and readers. On Summary and Structured, whose utility
is already near zero, the factorial excludes writer and same-family effects above $0.038$
(\supp{sec:readerfull}).

\subsection{What the instrument did not find}
\label{sec:r0}
\begin{figure}[t]
\centering
\includegraphics[width=\columnwidth]{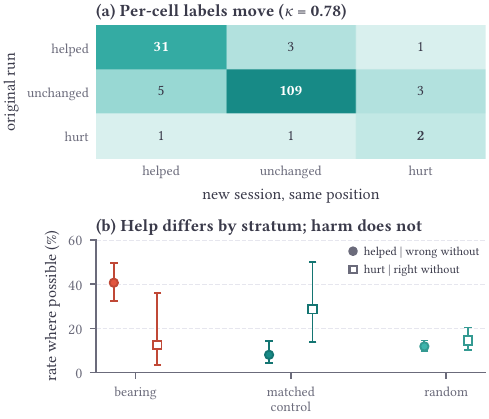}
\caption{What single cells can and cannot show (exploratory).
\textbf{(a)}~Per-cell labels (helped, unchanged, hurt) in the original run
against a new session at the same memory position ($\soStabSameN$ replicate
cells; diagonal agrees). At a new position $\kappa$ falls to $\soStabDiffK$, and
between two readers it is $\raReaderKappa$.
\textbf{(b)}~Help among cells the reader got wrong without the memory (filled)
and harm among cells it got right (hollow), by stratum, Wilson 95\% intervals:
help differs sharply, harm does not detectably.}
\label{fig:cells}
\Description{Two panels. Top: Cohen's kappa of 0.78, 0.65 and 0.58 for a new session at the same position, a new session at a new position, and a second reader, with 91, 87 and 82 percent raw agreement. Bottom: help rates of about 41 percent for bearing against 8 and 12 percent for control and random, while harm rates overlap across strata.}
\end{figure}
The study was designed to measure sign reversal: one memory helping one task
and harming another. Two naive comparisons mislead. Flip rates depend on cells
per memory ($\raCellsPerMemBear$ for \textsc{bearing}, $\raCellsPerMemRand$ for
\textsc{random}): at matched counts and no-memory outcomes, \textsc{random} also
shows no flip in $\lfCondFlipPZero\%$ of resamples. And harm shares depend on
baseline accuracy, since a memory can only hurt where the reader was right
without it. Conditioning on that opportunity (Figure~\ref{fig:cells}b;
\supp{sec:negative}), answer-bearing
pairs hurt in $\raHurtRateBear\%$ $[\raHurtLoBear, \raHurtHiBear]$ of the
$\raHurtOppBear$ baseline-correct pairs where they could, random-task pairs in
$\raHurtRateRand\%$ $[\raHurtLoRand, \raHurtHiRand]$: \emph{we detect no excess
harm}, but with 16 pairs the design could detect only a bearing harm rate of
$\rvHarmMdePct\%$ or more, so this does not establish the absence of harm. Matched-control pairs hurt in
$\raHurtRateCtrl\%$ $[\raHurtLoCtrl, \raHurtHiCtrl]$, an interval that overlaps
random's. The strata
differ in help: $\raHelpRateBear\%$ $[\raHelpLoBear, \raHelpHiBear]$ of the
cells where the reader was wrong, against $\raHelpRateRand\%$ for
\textsc{random}. The planned variance decomposition is not identifiable in the main run
(exploratory estimate in \supp{sec:sophia}).

\section{Discussion and Limitations}
\label{sec:discussion}
\label{sec:limitations}

Our measurements inform two memory-manager decisions without implementing
either: what to keep after an episode and what to show at each later step.
\emph{Retention} would require combining task-conditioned utility profiles with
an anticipated future-task distribution and memory costs; our measurements
characterize a defined evaluation task set, not a deployed agent's future
return. \emph{Exposure} is the standard retrieval problem: generic retrieval
recovers screen-defined answer-bearing memories far above chance but usually
not in its top ten (Section~\ref{sec:similarity}), so a reliable locator remains
open. For representation, the evaluated compressed representations retain less
answer-bearing text and gain less than Raw, including when total context is
fixed.

\paragraph{The pairing is constructed.}
The memories are real trajectory spans, but their pairing to tasks is a seeded
rule fixed before measurement, so every rate describes a defined evaluation task set rather than naturally
occurring future tasks. The observed harm rates should therefore not be
interpreted as population rates for deployed memory systems.

\begin{figure}[t]
\centering
\includegraphics[width=\columnwidth]{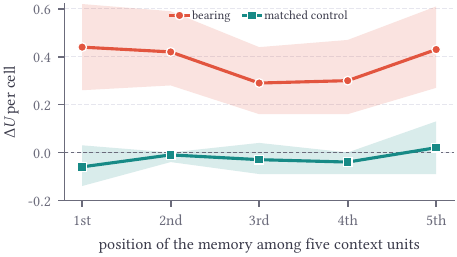}
\caption{Position affects per-pair measurements; the aggregate bearing--control
contrast remains positive after position standardization (exploratory). Per-cell $\Delta U$ by the memory's position among the five
context units in the second-session run, with task-clustered 95\% bands: the
bearing effect dips in the middle and recovers at the end, while the control
stays near zero at every position.}
\label{fig:position}
\Description{A line chart over five positions. The bearing line is about 0.44 and 0.42 at the first two positions, about 0.29 and 0.30 in the middle, and 0.43 at the last, with a wide band; the control line stays near zero throughout.}
\end{figure}

\paragraph{Session and position variance.}
Re-issuing $\lfDriftN$ byte-identical no-memory requests a day later changed
$\lfDriftReward$ rewards, so we compare only arms executed in one session. We then
measured both sources, session and position, directly by re-running the headline cells in a new session
with position redrawn per replicate (Section~\ref{sec:screened}). The aggregate
contrast keeps its direction. Single cells move (Figure~\ref{fig:cells}a). Position alone changes
$\soPosChgPct\%$ of labels and accounts for $\soPosVarPct\%$ of the variance of
$\Delta U$, and the bearing effect is weaker in the middle of the context
($\soPosMid$) than at its ends ($\soPosEdge$; Figure~\ref{fig:position}),
echoing~\cite{lostmiddle}. Labels agree with the original run at
$\kappa = \soStabSameK$ at the same position and $\soStabDiffK$ at a new one, and
changes almost always move to or from \emph{unchanged} (Figure~\ref{fig:cells}a). Standardizing the published contrast over position leaves it at
$\lfScrPosStd$.

\paragraph{Scope.}
The main results concern a single-call factual-recall reader evaluated on
the web and enterprise domains of LongMemEval-V2. The experiments use one
serving cluster and two reader checkpoints, limiting generalization to
other model families and deployments. The software-engineering and
scientific-coding evaluations show no measurable memory effect; the
positive factual-recall finding should not be extended to those settings. Single-item insertion cannot see
redundancy or complementarity between units, which Shapley-style attribution
could~\cite{meclear}. Our filler
comes from trajectories that host no bearing memory, so it never contains
competing evidence; realistic retrieval would, which likely makes our bearing
effect larger than a deployed system would see. Our leakage-audit rates are lower bounds (\supp{sec:extlimits}).

\paragraph{Release and reproducibility.}
\label{sec:release}
The Conditional Memory Utility Dataset is publicly available on
\href{https://huggingface.co/datasets/AgentsSci/AAMAS2027-ConditionalMemory}{Hugging Face}.
The frozen \texttt{v0.6-}\allowbreak\texttt{length-control} snapshot contains 16,816 evaluation
records, with paired outcomes, costs and available text, plus schema and
validation code. The data card documents the included conditions, missing
records and text withheld under upstream restrictions. This versioned snapshot
does not cover every later manuscript analysis, including the fixed-total-context
experiment. Per-pair labels depend on the evaluation condition; they are
observations, not unconditional targets for training a memory gate.
See \supp{sec:repro} for coverage and reuse details.

\paragraph{Conclusion.}
Evaluate memory compression by the task utility it preserves alongside its
costs. On LongMemEval-V2 factual-recall tasks, answer-bearing pairs gain more
than matched controls, and the tested summaries and structured records gain
less than Raw even at fixed total context, consistent with reduced retention
of relevant content. Our paired protocol makes these effects measurable
without exposing the future task to the memory writer. The headline contrast
stays positive across the tested sessions, positions and readers, while
individual labels remain condition-dependent. These factual-recall gains do
not extend to the tested coding benchmarks.

\section*{Use of AI-Assisted Technologies}
We follow the stronger clause of the AAMAS policy. AI assistants helped with the literature review, coding, analysis, and paper writing under human guidance. The authors made the research decisions, interpreted the results, verified all numbers, and checked all citations.

\begin{acks}
This research used resources of the Argonne Leadership Computing Facility, a U.S. Department of Energy (DOE) Office of Science user facility at Argonne National Laboratory (ANL) operated under Contract No.~DE-AC02-06CH11357. The work was also supported under the same contract by the DOE Office of Science's Advanced Scientific Computing Research Program and by Laboratory Directed Research and Development (LDRD) funding from ANL, provided by the Director, DOE Office of Science. The work of the University of Houston coauthors was supported in part by the National Science Foundation under grants CAREER-2305491 and CNS-2504762.
\end{acks}

\bibliographystyle{ACM-Reference-Format}
\bibliography{p1refs}

\clearpage

\clearpage
\section*{Supplementary Material}
\label{sec:supplementary-material}

\setcounter{section}{0}
\setcounter{subsection}{0}
\setcounter{figure}{0}
\setcounter{table}{0}
\renewcommand{\thesection}{S\arabic{section}}
\renewcommand{\thesubsection}{\thesection.\arabic{subsection}}
\renewcommand{\thefigure}{S\arabic{figure}}
\renewcommand{\thetable}{S\arabic{table}}

\providecommand{\main}[1]{main paper, Section~\ref*{M-#1}}

\noindent\textbf{Guide to this supplement.}
The paper is self-contained, and this document holds the detail behind it. Section~\ref{sec:protocol} gives the full protocol: how the
writer is kept blind to future tasks, how target position is controlled,
both gold-free screens, and the leakage audit.
Section~\ref{sec:benchdetails} describes the benchmark adapters, the scorer,
pairing and power, and why each secondary benchmark has the yield it has.
Section~\ref{sec:config} lists the frozen execution configuration, the noise
floors, the label definitions, and why the additive null and the variance
decomposition matter. Section~\ref{sec:lmresults} adds LongMemEval-V2
results: the cell-level sparsity map, the reader capability check, the
length-control design, and the formation-cost sensitivity. Section~\ref{sec:readerfull} reports the
second-reader study and the writer-by-reader factorial in full, and
Section~\ref{sec:crossbenchtable} gives the cross-benchmark contrasts as a
table. Section~\ref{sec:negative} reports the sign-reversal analysis the study
was designed around, including why naive sign comparisons mislead and the
transition analysis that replaces them. Section~\ref{sec:repro} covers data, code and
prompt availability, licensing and reproducibility, and
Section~\ref{sec:extlimits} the limitations not needed to read the main result. Section~\ref{sec:prereg} is the pre-registration record: every
pre-registered analysis, its anchor, and every recorded deviation. Section~\ref{sec:sophia} reports three follow-up runs: a second session
with randomized position, a fixed total context, and a retrieval evaluation. Every table and figure states its
benchmark, population and denominator.

\section{Protocol details}
\label{sec:protocol}

\subsection{Keeping the writer blind to future tasks}
\label{sec:taskblind}
Memory formation must be a function of the past episode alone, $m = W(e)$.
If the writer could see a future task, a positive $\Delta U$ would show only
that a model can be told the answer and copy it forward. We enforce this with
four structural barriers rather than by convention. (1)~\emph{Arity}: the
writer function takes a segment and nothing else. (2)~\emph{A closed field
set}: the writer emits only the fields of its schema. (3)~\emph{A metadata
denylist}: the weakest barrier, present to make a plumbing mistake fail loudly.
(4)~\emph{A single interpolation point}: the prompt has one placeholder, the
segment text. Each barrier carries a negative control that plants the defect
and requires the check to fire. Prompt identity is derived from the prompt
file's bytes, not from a hand-maintained version constant that could report a
stale version.

\subsection{Target position is held fixed}
\label{sec:position}
The \textsc{with} arm holds one more item than the \textsc{without} arm, so
some rule must place it, and that rule is not neutral. Holding the memory,
question, filler, model, decoding and seed fixed and varying only the
position of the target memory changes the answer in 8 of 12 probe cells. Measured at scale in the second-session
run (Section~\ref{sec:sophia}), position changes $\soPosChgPct\%$ of labels. A
comparison that does not control position therefore measures position
together with presence. We hold position fixed across replicates, vary it
across cells, and record it on every row. This is also what makes the
second-reader comparison (\main{sec:readertransfer}) a comparison of readers
and not of positions. The probe is small ($n{=}12$), and we treat the rate
as provisional.

\subsection{The bearing screen and both gold-free screens}
\label{sec:goldfree}
The bearing screen reads the memory text, the task's gold answer, and the
task's evaluation specification; it never reads the question. A pair is
\textsc{bearing} only when the benchmark's scorer marks the memory text
correct against the gold answer \emph{and} the test fires on at most eight of
the 865--1{,}024 units in the task's domain pool. The second clause matters.
An earlier, non-discriminative version of the flag fired on nearly every
candidate: its rate on paired versus unpaired memories was in a ratio of
$0.995$, so it said nothing about any one of them.

We pre-registered two gold-free screens before execution. Each reads the
memory, the question and the same-domain pool, and none of the gold answer,
the reward, any model output, or the evaluation specification. We report
both, because reporting only the screen that worked would select an
instrument on its outcome.
\begin{itemize}
\item \emph{Model-based screen.} The judge is $\lfGfJudge$, the reader's own
model, at temperature 0, asked whether the memory contains the specific
information the question asks about. On cells the gold screen never
selected, its picks move the reader from $\gfWithoutPct\%$ correct without the
memory to $\gfWithPct\%$ with it ($\gfN$ rows, $\gfCells$ cells); its matched
control moves from $\gfCtlWithoutPct\%$ to $\gfCtlWithPct\%$ ($\gfCtlN$ rows,
$\lfGfCtlCells$ cells). The contrast of $\gfDiff$ has a task-clustered interval
of $[\lfGfTaskLo, \lfGfTaskHi]$ ($\lfGfTaskN$ tasks), an episode-clustered one
of $[\lfGfEpLo, \lfGfEpHi]$ ($\lfGfEpN$ episodes), and a paired-by-memory
difference of $\lfGfMemD$ over $\lfGfMemN$ memories. Against the gold screen it
recovers $\lfGfBoth$ of $\lfGfGold$ gold-bearing pairs ($\lfGfRec\%$ recall),
and $\lfGfBoth$ of its $\lfGfPicks$ picks are gold-bearing ($\lfGfPrec\%$). The
run is in one session, separate from the gold run.
\item \emph{Lexical screen.} It returns no effect. A diagnostic independent of
any utility explains why: its bearing cells fail in both arms at a rate of
$\gfLexBaf$, \emph{worse} than unscreened random pairing, against
$\gfLlmBaf$ for the model-based screen. A criterion whose selections are
harder than a random draw has not located bearing pairs.
\end{itemize}

\subsection{Leakage audit}
\label{sec:leakage}
The writer invariant excludes \emph{task-directed writing}, not a naturally
occurring memory that happens to contain a later answer. Such a memory is
genuinely useful, and removing it would condition the cell set on the
outcome. We therefore audit every cell rather than filter it. The detector
reuses the benchmark's own scorer and asks whether the rendered memory text
alone would score correct against the gold answer. For \textsc{bearing} cells
the answer is yes by construction, since that is half the bearing definition.
The audit carries information for the control and \textsc{random} strata and
for the Summary and Structured texts, where containment is not guaranteed and
can be conditioned on in analysis. It is validated by 17
controls (plants it must catch, clean cases it must not fire on, and two
pinned known false negatives) and aborts if any control fails.

\subsection{Three kinds of zero in the cost ledger}
\label{sec:zeros}
Raw makes zero writer calls; Summary and Structured make one each per
segment. The ledger distinguishes three zeros that are routinely merged: a
measured zero (Raw makes zero extraction calls), an absent measurement (no
call was made, so the cell leaves that denominator), and an unreported field
(the endpoint exposes no counter). Failed writes count: a write that failed
after consuming a call still cost that call.

\section{Benchmark adapters and coverage}
\label{sec:benchdetails}

\paragraph{LongMemEval-V2 haystacks.}
On the \texttt{small} tier each domain has exactly one haystack, identical
across questions even as an ordered tuple, so every memory is available to
every question in its domain and there is no availability confound between
arms. The same test returns 222 and 211 distinct haystacks on
\texttt{medium}, a positive control showing that it distinguishes the two
cases. Trajectory lengths were measured with the benchmark's own tokenizer
over 200 trajectories and 5{,}095 states.

\paragraph{Scorer dispatch.}
There are eight distinct evaluation-specification strings, not six: two
phrase-match variants differ only in their separator. The adapter therefore
dispatches on the full string, never on the function name, and calls the
benchmark's scorer verbatim. Its phrase matcher tests gold-phrase
\emph{containment}, not set equality, which a name-based reimplementation
gets wrong. An LLM-judged task raises an error rather than returning a
default; a silent \texttt{False} would turn ``cannot evaluate'' into ``wrong
answer'' for 156 of 451 questions in both arms.

\paragraph{Pairing and power.}
Seeded deterministic pairing within domain, stratified by the benchmark's own
question-type taxonomy, gives 1{,}889 memories $\times$ 8 tasks $=$ 15{,}112
cells, and every memory was verified to receive exactly 8. The build asserts
that the same seed reproduces the grid byte-identically \emph{and} that a
different seed changes it, so the reproducibility check is not vacuous. That
grid is the available pairing, not the executed design. The floor of 60
memories was fixed before running from a recovery study on synthetic data
with known planted components; the same study showed that widening the task
axis helps more than adding memories.

\paragraph{BEAM.}
BEAM~\cite{beam} was in our original plan as a multi-domain check but cannot
supply the task axis: 92.6\% of its cited turns are cited by exactly one
question. Being CC~BY-SA~4.0, it would also have to ship separately, since
ShareAlike would otherwise relicense the rest of the release.

\paragraph{SciCode.}
The memory unit is a reusable helper routine with its procedure, which
transfers across problems. Against an early feasibility estimate of 8--12
pairs, the full candidate pool measured on the built adapter is $\sciPairs$
donor$\to$recipient pairs ($\sciMems$ memories, $\sciTasks$ tasks), and we
ran all of it.

\paragraph{SWE-bench Verified.}
Pairing is plentiful but the \emph{measurable} yield is low. Across
$\covSwRepos$ repositories, after a localizer that misses the edit region on
$\sweLocMiss$ cells and edits that neither apply nor produce a test verdict,
$\sweMatched$ cells are interpretable under all three representations, and
$\sweDomRepoPct\%$ come from one repository. A scoring defect was corrected before these counts: a test suite
that printed no verdict at all (an import or collection error) had been
recorded as a reward of zero. Those $\lfSweDemoted$ arm-reads
($\lfSweDemDjango$ django, $\lfSweDemSympy$ sympy) are now unscorable, and
every contrast includes zero after the correction. The matched set of
$\lfSweMatched$ cells is pooled from four runs across the two repositories; it grew as runs were added, not through the correction, which can only
remove cells. Table~\ref*{M-tab:coverage} of the main paper counts SWE-bench in
(memory, task, representation) cells; counts per recipient task (107/37/70)
describe a different unit.

\section{Execution configuration and statistics}
\label{sec:config}

\paragraph{Frozen configuration.}
Cluster and serving configuration are part of a cell's experimental
identity, and every row records the full pinned set: reader and writer
\texttt{gpt-oss-120b} on one cluster, low reasoning effort,
temperature 0, \texttt{max\_tokens} 512, 16 concurrent workers,
SFR-Embedding-\allowbreak Mistral at dimension 4096, and the benchmark's own tokenizer. A
change to any of them defines a new arm rather than additional data. Three
settings were fixed by measurement. An empty or truncated completion raises
and the cell is re-run, since scoring it as zero would push $\Delta U$
negative and manufacture harmful memories. Concurrency sits at a measured
saturation ceiling, with the ladder pushed past the chosen value. Responses
are cached on all twelve fields of a cell's identity, verified by varying one
field and confirming a cache miss. The embedding model's determinism was
checked directly: the cosine of identical texts is exactly 1.0.

\paragraph{Noise floors.}
\label{sec:floors}
$F_{\mathrm{exec}}$ is outcome variation across seeds with everything else
fixed, measured per benchmark with a bootstrap interval and within a single
execution session. $F_{\mathrm{empty}}$ is the variation of the no-memory
arm, available at no extra cost because that arm is identical for every
memory paired with a task. $F_{\mathrm{judge}}$ is scorer disagreement on
byte-identical outputs, from re-scoring stored generations; we store full
output text rather than a digest so that this re-scoring is possible. The
floor is defined on the $\Delta U$ scale, since
$\mathrm{var}(\Delta U) = \mathrm{var}_{\mathrm{with}} +
\mathrm{var}_{\mathrm{without}} - 2\,\mathrm{cov}$; per-arm floors are
reported alongside. One hazard would fabricate the whole result: the
\textsc{without} arm's inputs are identical across memories, so a cache
serving it would reuse one generation everywhere and let every $\Delta U$
clear a zero floor. That arm is re-generated per (task, replicate) rather than served from the
response cache; in the screened run all $\lfScrRealCalls$ calls reached the
endpoint, with $\lfScrCacheHits$ cache hits.

\paragraph{Labels and flips.}
A cell is robust useful when $\mathrm{CI}_{\mathrm{lower}} > 0$ and
$\overline{\Delta U} > F$, robust harmful when $\mathrm{CI}_{\mathrm{upper}}
< 0$ and $|\overline{\Delta U}| > F$, and neutral otherwise, so a cell can be
statistically non-zero and still neutral. A memory shows a robust sign flip
only if two of its cells independently meet the useful and harmful criteria.
Every flip rate carries its denominator (memories on at least two tasks) and
a Wilson interval beside the naive rate, and a cell needs at least two
replicates to enter one. Because replicates agree on every primary cell, a
robust label reduces in practice to the sign of $\Delta U$, and flip rates are
comparable only at matched cells per memory (Section~\ref{sec:negative}).

\paragraph{The additive null.}
A flip is a property of $\Delta U$'s sign; interaction is a property of its
variance. Under $\Delta U(m,t) = \mu + a_m + b_t$ with no interaction, a
memory still helps one task and harms another whenever $b_t$ crosses
$-(\mu + a_m)$. Simulating that null over 20 trials of 200 memories $\times$
8 tasks gives flip rates of 53.5\%--92.0\% across five settings of $\mu$ and
the memory and task standard deviations. A high flip rate is therefore
consistent with memory value being a scalar plus a task-difficulty term.

\paragraph{The variance decomposition.}
We pre-registered a decomposition $\Delta U(m,t,s) = \mu + a_m + b_t +
(ab)_{mt} + \varepsilon_s$ into component variance shares against
$F_{\mathrm{exec}}$, with a method-of-moments fallback fixed in advance. We
recorded the structural precondition before executing: components are
identified only relative to a residual scale. It fails here because the
reader is deterministic within a session, so we quote no share
(Section~\ref{sec:negative}). Standard errors are clustered by task
throughout.

\paragraph{Similarity and cost.}
Semantic distance is the cosine of the pinned embeddings of memory and task,
banded by terciles fixed before outcomes were examined, since cosine scales
are not comparable across corpora. The seven cost families are
\textbf{storage} (memory tokens held), \textbf{retrieval} (latency, tokens
returned), \textbf{exposure} (tokens placed in the reader's context),
\textbf{generation} (latency, input, output and reasoning tokens),
\textbf{extraction} (formation cost), \textbf{API and compute}, and
\textbf{selectivity} (how much of what was retrieved was worth exposing).
Exposure is recorded separately from retrieval because the gap between them
is truncation loss.

\section{Additional LongMemEval-V2 results}
\label{sec:lmresults}

\subsection{The cell-level sparsity map}
\label{sec:sparsitymap}
\begin{figure}[t]
\centering
\begin{tikzpicture}[
  font=\scriptsize,
  tk/.style={font=\tiny, text=black!55},
  hd/.style={font=\tiny\bfseries, text=black!80},
  lg/.style={font=\tiny, text=black!60},
  br/.style={draw=black!45, line width=0.35pt},
]
\def\cw{3.5pt}\def\ch{2.1pt}
\foreach \x/\y in \spZero
  { \fill[teallight!18] (\x pt,\y pt) rectangle +(\cw,\ch); }
\foreach \x/\y in \spNeg
  { \fill[coral] (\x pt,\y pt) rectangle +(\cw,\ch); }
\foreach \x/\y in \spPos
  { \fill[tealdeep] (\x pt,\y pt) rectangle +(\cw,\ch); }
\foreach \mi/\nm/\rt/\lo/\hi in {%
    \spBearMid/{\textsc{bearing}}/\spBearRate/\spBearLo/\spBearHi,
    \spCtrlMid/{matched control}/\spCtrlRate/\spCtrlLo/\spCtrlHi,
    \spRandMid/{\textsc{random}}/\spRandRate/\spRandLo/\spRandHi}{
  \node[tk, text=black!75, anchor=south] at (\mi,7pt) {\nm};
  \node[tk, anchor=north] at (\mi,{-\spTotH-1pt})
    {\rt\%\ \,{\tiny[\lo, \hi]}};
}
\node[tk, anchor=north] at (\spBearMid,{-\spTotH-9pt}) {\spBearPos/\spBearN};
\node[tk, anchor=north] at (\spCtrlMid,{-\spTotH-9pt}) {\spCtrlPos/\spCtrlN};
\node[tk, anchor=north] at (\spRandMid,{-\spTotH-9pt}) {\spRandPos/\spRandN};
\node[tk, anchor=north] at (78pt,{-\spTotH-17pt})
  {cells where the memory helped, with Wilson 95\% interval};
\draw[br] (-3pt,0pt) -- (-3pt,{-\spTotH});
\node[tk, rotate=90, anchor=south] at (-8pt,{-0.5*\spTotH})
  {\spMems\ memories};
\fill[tealdeep] (0pt,20pt) rectangle +(\cw,\ch);
\node[lg, anchor=west] at (5pt,21pt) {helped};
\fill[coral] (30pt,20pt) rectangle +(\cw,\ch);
\node[lg, anchor=west] at (35pt,21pt) {hurt};
\fill[teallight!18] (54pt,20pt) rectangle +(\cw,\ch);
\node[lg, anchor=west] at (59pt,21pt) {no change};
\node[lg, anchor=west] at (95pt,21pt) {(blank $=$ not run)};
\end{tikzpicture}
\caption{\textbf{Utility is sparse over future tasks, and the sparsity is
structured.} One square per \emph{measured} cell --- all $\spCells$ of them,
$\spMems$ memories $\times$ their own future tasks, one representation and
one frozen reader. $\Delta U$ takes exactly three values here, so the marks
are discrete rather than shaded. Only $\spNonzero$ cells move at all:
$\spZeroPct\%$ of measured cells are an exact zero, which is the quantity a
single global memory score has to summarize. The three blocks are the same $\spMems$
memories against bearing, matched-control and unscreened tasks, so the
comparison is paired within memory. Rows are ordered by \textsc{bearing}
count; that arranges the rows but cannot create the block contrast, which is
$\spBearRate\%$ against $\spRandRate\%$ with disjoint Wilson intervals. A row's squares are that memory's own tasks in its
own order, so a \emph{column is not a task}: the blocks span $\spBearTasks$,
$\spCtrlTasks$ and $\spRandTasks$ distinct tasks against widths of 9, 9 and
15. Unrun cells are blank rather than padded --- the rows are ragged, and
padding them would spread the same marks over more area and collapse the
contrast to about $1\times$.}
\label{fig:sparsitymap}
\end{figure}
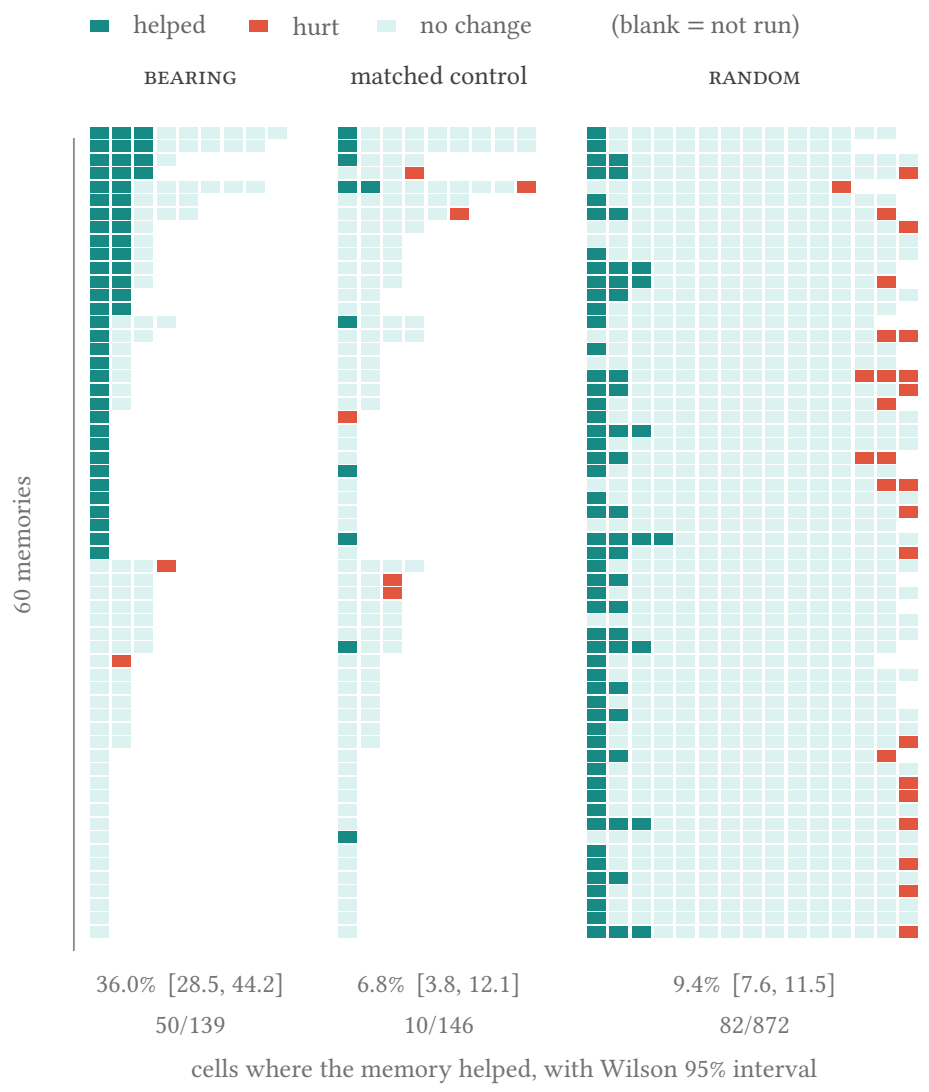

Figure~\ref{fig:sparsitymap} draws every one of the $\spCells$ measured cells
of the screened design. It is the cell-level view behind
Figure~\ref{fig:overview}b and Table~\ref*{M-tab:screened} of the main paper.
$\Delta U$ takes exactly three values here, because the reward is binary and
both replicates agree on every cell, so the marks are discrete rather than
shaded. Unmeasured slots are left blank: rows are ragged (at most 9, 9 and 15
cells per memory in the three blocks), and padding them to a rectangle would
spread the same marks over more area. Per drawn square the \textsc{bearing}
rate is $\spBearRate\%$ against $\spRandRate\%$ for \textsc{random}; per padded
area the two would look the same, which is why the figure does not pad.

\subsection{Embedding similarity in detail}
\label{sec:simdetail}
This section holds the detail behind \main{sec:similarity}. Cosine similarity between memory and task, under the pinned embedding model,
is nearly uninformative about $\Delta U$: Spearman $\rho=\simRhoAll$ over all
$\simCells$ cells and $\rho=\simRhoRand$ on unscreened cells alone. The pooled series rises slightly at high
similarity partly because \textsc{bearing} cells concentrate there, and within
\textsc{bearing}, $\rho = +0.164$.

This result is narrower than it looks, for two reasons. First, 96.6\% of the
similarities are computed from a truncated prefix of the memory, and bearing
content sits mostly near the end of a segment: the first $N$ tokens retain it
in 0.7\% of cells against 28.9\% for the last $N$ (\main{sec:repr}).
The embedding therefore mostly never saw the content that matters. Second,
Spearman $\rho$ over an outcome that is exactly zero in $\spZeroPct\%$ of cells
is depressed by ties and is not what a retriever needs, which is to rank the
bearing memory first. We therefore claim only that this embedding
configuration does not track utility, not that embedding similarity cannot.

\subsection{The reader can use the memory}
\label{sec:needle}
We planted a per-cell unique code inside the target memory, in the real
$\sim$27k-token context, and asked the reader for it. The base reader
recovers it in 22 of 22 cells and in 0 of 17 when nothing is planted. Two
larger models also recover 22 of 22 and solve no cell the base reader
missed. The reader can recover planted content at full context length. This makes
a general inability to access the context less likely, but does not rule
out difficulty locating or reasoning over natural evidence. In the two
earlier unscreened arms, accuracy also rises monotonically with how
discriminatively the memory matches the task: $0.1429$, $0.4545$, $0.5000$.

\paragraph{Earlier unscreened arms.}
The same contrast appears in two earlier, unscreened prompt arms, where
bearing was identified afterwards from content alone. On the
$\lfUnsHwN$ and $\lfUnsOffN$ bearing cells of the two arms, mean $\Delta U$ is
$+0.0909$ and $+0.1818$ (2 and 4 net helped cells), against $-0.0066$ and
$-0.0153$ elsewhere; pooled, the difference is $+0.1473$ (permutation
$p = 0.0029$ over 20{,}000 shuffles). These are small samples, and we read them
only as agreeing in sign with the screened design.

\subsection{Length control and budget sweep}
\label{sec:lengthcontrol}
\begin{figure}[t]
\centering
\includegraphics[width=\columnwidth]{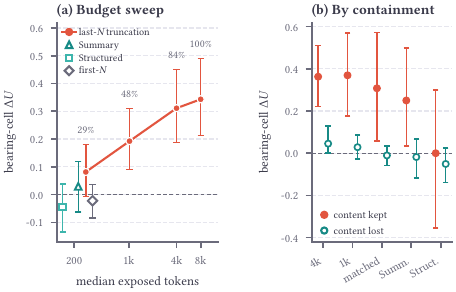}
\caption{Length-control run on LongMemEval-V2: $\Delta U$ on the
$\lcBearCells$ bearing cells, task-clustered 95\% intervals.
\textbf{(a)}~Raw truncated to its last 4,096 and 1,024 tokens and to each
cell's Summary length (line), against Summary, Structured and the first-$N$
window at that length; percentages give the share of cells whose text still
contains the bearing content. \textbf{(b)}~The same arms split by whether
their text kept the bearing content (filled) or lost it (hollow).}
\label{fig:lenctl}
\Description{Left: bearing-cell utility rising with exposed tokens from about 0.08 at 280 tokens to 0.34 at 8,000, with Summary, Structured and first-N near zero. Right: for each arm, utility near 0.25 to 0.37 when the bearing content is kept and near zero when it is lost, except Structured whose kept estimate is near zero with a wide interval.}
\end{figure}

The length control was frozen before execution. Its extractive arm keeps the
final $N$ tokens of the same source segment, with $N$ set per cell to that
cell's own Summary token count, so it matches Summary's budget without any
model choosing content; a second arm keeps the first $N$ tokens instead, and
last-$N$ versus first-$N$ was registered as the decisive contrast. All three
pre-registered contrasts are computed on the $\lcBearCells$ bearing cells
(two replicate rows each, $\lcPrimaryN$ rows) and clustered over
$\lcBearTasks$ tasks: extractive versus Summary ($\lcPrimary$
$[\lcPrimaryLo, \lcPrimaryHi]$), Raw versus matched extractive ($\lcBudget$
$[\lcBudgetLo, \lcBudgetHi]$), and last-$N$ versus first-$N$ ($\lcEnds$
$[\lcEndsLo, \lcEndsHi]$).

The same run truncated Raw to its last 4,096, 1,024 and 20,000 tokens; the last
equals Raw, since no segment is longer. Table~\ref{tab:sweep} gives each arm's
bearing-cell $\Delta U$, its retention of the bearing content, and $\Delta U$
split by retention. \texttt{make\_lenctl.py} computes every entry from the row
file, whose checksum it verifies against the run manifest, and first checks
that it reproduces the three pre-registered contrasts. The run is not strictly
one session: $\lfLcCacheHits$ responses came from a cache filled minutes
earlier, and $\lcNullRows$ rows with a null reward are excluded, never zeroed.

\begin{table}[t]
\centering\footnotesize
\setlength{\tabcolsep}{2.4pt}
\caption{Length-control run on LongMemEval-V2: bearing-cell $\Delta U$ by arm
($\lcBearCells$ cells, task-clustered 95\% intervals), share of cells whose text
still contains the bearing content, and $\Delta U$ split by whether it does.
Exploratory beyond the three pre-registered contrasts.}
\label{tab:sweep}
\begin{tabular}{@{}lrrrrr@{}}
\toprule
arm & median tokens & kept & $\Delta U$ & $\Delta U\mid$kept & $\Delta U\mid$lost \\
\midrule
Raw (full) & \lcRawTok & \lcRawKept\% & \lcRawM\ [\lcRawLo, \lcRawHi] & \lcRawKM & -- \\
last 4,096 & \lcFourTok & \lcFourKept\% & \lcFourM\ [\lcFourLo, \lcFourHi] & \lcFourKM & \lcFourLM \\
last 1,024 & \lcOneTok & \lcOneKept\% & \lcOneM\ [\lcOneLo, \lcOneHi] & \lcOneKM & \lcOneLM \\
last $N$ (matched) & \lcMatchTok & \lcMatchKept\% & \lcMatchM\ [\lcMatchLo, \lcMatchHi] & \lcMatchKM & \lcMatchLM \\
Summary & \lcSumTok & \lcSumKept\% & \lcSumM\ [\lcSumLo, \lcSumHi] & \lcSumKM & \lcSumLM \\
Structured & \lcStrTok & \lcStrKept\% & \lcStrM\ [\lcStrLo, \lcStrHi] & \lcStrKM & \lcStrLM \\
first $N$ (matched) & \lcFirstTok & \lcFirstKept\% & \lcFirstM\ [\lcFirstLo, \lcFirstHi] & -- & \lcFirstLM \\
\bottomrule
\end{tabular}
\end{table}

\paragraph{Is the kept subset easier?} Retention is a property of the text,
fixed before any outcome, but the cells where a compressed text keeps the
content could still be easier ones. They are not for Summary or the matched
truncation: on the same cells, Raw gains $\lcSelSumKept$ where Summary kept the
content and $\lcSelSumLost$ where it lost it ($\lcSelMatchKept$ and
$\lcSelMatchLost$ for the matched truncation). For Structured the kept cells are
\emph{harder} ($\lcSelStrKept$ against $\lcSelStrLost$), which accounts for
its low kept estimate.

\paragraph{Position.} Target position is identical across all eight arms
within each cell (asserted). Between strata its distribution differs mildly
($\chi^2$ $p = \lcPosChiP$); the position-standardized Raw
\textsc{bearing}$-$control contrast is $\lcPosStd$ against $\lcPosUnadj$
unadjusted, and positive at every position.

\subsection{Formation-cost sensitivity}
\label{sec:reuse}
\begin{figure}[t]
\centering
\includegraphics[width=\columnwidth]{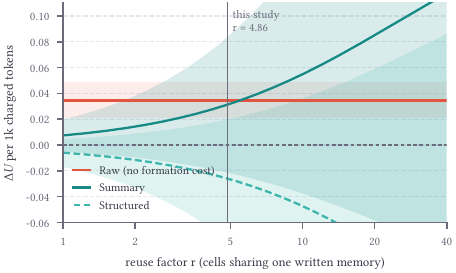}
\caption{Exposure efficiency ($\Delta U$ per thousand charged tokens) against
the reuse factor, on the $\frCells$ representation cells of LongMemEval-V2.
Lines use each representation's paired \textsc{bearing}$-$control effect;
bands use its 95\% interval over memories. Raw has no writer call and does not
depend on $r$. The vertical line marks the reuse factor of this study.}
\label{fig:reuse}
\Description{Line chart of efficiency against reuse factor on a log axis: Raw is flat near 0.034, Summary rises and crosses Raw near r of 5.5 with a wide band spanning zero, Structured is below zero.}
\end{figure}

Formation cost per cell is the writer's tokens divided by the reuse factor
$r$, the number of cells that share one written memory. Our design fixes
$r = \frReuse$, so we recompute exposure efficiency over $r$ from the same
utilities and token counts (Figure~\ref{fig:reuse}). One Summary costs about
$\raFormSum$ writer tokens and one Structured record about $\raFormStr$. On
point estimates Summary overtakes Raw only for $r \gtrsim \raReuseStar$, but
its utility interval spans zero at every $r$, so the crossing point is not
estimated with any precision. Under the sparsity we measure, a memory that
bears on one task may be shown for few others, which pushes $r$ down and
formation cost up. The relevant factor in deployment is exposures per write,
not cells per memory. Input and output tokens are priced identically and prompt
caching is ignored; since output tokens usually cost more, both choices
understate the writer's cost.

\section{Second reader and writer, $\times$, reader factorial}
\label{sec:readerfull}

\paragraph{Design.}
The second-reader study was pre-registered before any design-cell call was
issued, with no transfer, the same ordering at a different magnitude, and a
different representation preference all committed as valid outcomes. The
memory file, prompt, filler pool, every rendered request, and each cell's
target position were hash-asserted against the first run at call time, and
the text re-verified byte-identical on $\rgRows$ rows; in the session-matched
run all $\smIdentK$ of $\smIdentN$ joinable rows are hash-identical.

\paragraph{Session-matched result.}
Both readers ran in one session. Over $\rgSmCells$ matched Raw cells on
$\rgSmTasks$ tasks ($\rgSmN$ replicate rows; these three counts are different
units), mean $\Delta U$ is $\rgSmOss$ under the first reader and $\rgSmGem$
under the second, a paired change of $\rgSmDiff$ $[\rgSmLo, \rgSmHi]$. The
two readers' own 95\% intervals overlap; plotting those would invite a ``no
difference'' reading that the paired design rules out. The within-memory
\textsc{bearing}$-$control contrast is $\rgBmcRawOss$ $[\rgBmcRawOssLo,
\rgBmcRawOssHi]$ under the first reader and $\rgBmcRawGem$ $[\rgBmcRawGemLo,
\rgBmcRawGemHi]$ under the second ($\rgBmcRawOssMem$ memories). The no-memory
arm moves by $\rgWithoutDelta$, so part of the gain is a weaker reader rather
than a better memory.

\paragraph{Cross-session first execution.}
A first execution with the two readers a day apart returned $\rgRawDiff$
$[\rgRawLo, \rgRawHi]$, inside the session-matched interval. Re-run together,
that difference splits into $-0.0327$ attributable to the session and
$-0.0364$ to the reader, which halves the apparent reader share. We quote only
the session-matched estimate.

\paragraph{Cell-level agreement, session-matched run.}
These are the numbers behind Figure~\ref{fig:overview}f. Over the
$\rgSmCells$ matched Raw cells, the readers' three-way labels (helped,
unchanged, hurt) agree on $\raReaderLabAgree$ of $\rgSmCells$ cells, at Cohen's
$\kappa = \raReaderKappa$; $\smCZeroZero$ cells are unchanged under both and
can only agree. Of the $\raReaderMoving$ cells that move under either reader,
both mark $\smAgree$ as helped, and signs agree on $\smAgree$ of the $\smBoth$
cells that move under both. The readers agree on no harmed cell (the first
reader's $\raHurtOss$ and the second's $\raHurtGem$ do not overlap), but so few cells
are harmed that this cannot show whether harm labels reproduce.

\paragraph{Cell-level agreement, earlier cross-session run.}
The first execution, with the two readers a day apart, covers a slightly
different matched set ($\rgRawN$ cells) and is the only source for Summary and
Structured. There the readers' per-cell $\Delta U$ correlate at
$r{=}\rgRawRAll$ over all cells, of which $\rgRawZeroBoth$ are zero under both;
over the $\rgRawNNz$ cells non-zero under either reader, $r = \rgRawRNz$
($\rgSumRNz$ for Summary, $\rgStrRNz$ for Structured). Signs agree on
$\rgRawSignK$ of $\rgRawSignN$ Raw cells where both move, and labels on
$\rgRawLabK$ of $\rgRawLabN$. Summary $\rgSumDiff$ $[\rgSumLo, \rgSumHi]$ and
Structured $\rgStrDiff$ $[\rgStrLo, \rgStrHi]$ span zero with only
$\rgSumNNz$ and $\rgStrNNz$ cells non-zero under either reader. Because this
run spans sessions, its numbers mix reader and session differences, and they
should not be pooled with the session-matched ones. Costs are not pooled
across readers: the second reports no reasoning-token counts.

\paragraph{Writer, $\times$, reader factorial.}
\label{sec:factorial}
All four writer--reader arms ran in one session. Every one of the eight
contrasts (writer, reader, same-family and interaction, on Summary with
$n{=}274$ and Structured with $n{=}268$) spans zero. The intervals are
informative rather than underpowered: they exclude writer and same-family
effects larger than $0.0380$, against the $\rgSmDiff$ reader effect the same
apparatus detects. Raw has no writer and is not in the factorial, and two
checkpoints form a $2\times2$ design, not a population of models.

\section{Cross-benchmark contrasts}
\label{sec:crossbenchtable}
The one quantity all three benchmarks define identically is the per-cell
representation contrast: a representation's $\Delta U$ minus Raw's on the same
memory and task. It separates from zero on LongMemEval-V2 alone
(Figure~\ref{fig:overview}e): Summary$-$Raw is $\xbSRLmM$
$[\xbSRLmLo, \xbSRLmHi]$ and Structured$-$Raw is $\xbTRLmM$
$[\xbTRLmLo, \xbTRLmHi]$ on $\xbSRLmN$ cells. On SciCode ($n{=}\xbSRScN$) and
SWE-bench Verified ($n{=}\xbSRSwN$) every interval includes zero, but there the
memories themselves have no effect even in raw form: Raw's own $\Delta U$ is
$\rvSciRaw$ $[\rvSciRawLo, \rvSciRawHi]$ on SciCode, where $\rvSciRawTied$ of
$\rvSciRawN$ cells are unchanged, and $\rvSweRaw$ $[\rvSweRawLo, \rvSweRawHi]$
on SWE-bench. With no Raw effect to preserve or lose, a representation contrast
there is uninformative about representation; these benchmarks bound where
memory matters for this reader, not how it should be written
(Section~\ref{sec:crossbenchtable}).

Table~\ref{tab:crossbench} gives the values plotted in
Figure~\ref{fig:overview}e. We compare contrasts rather than
per-representation $\Delta U$ because each benchmark carries its own
nondeterminism. The no-memory arm, whose input is identical across
representations, still disagrees with itself on
$\xbFloorLmK/\xbFloorLmN{=}\xbFloorLm$ of LongMemEval-V2 cells and
$\xbFloorScK/\xbFloorScN{=}\xbFloorSc$ of SciCode cells, and a paired contrast
removes that term. SWE-bench has no such floor because its no-memory arm is
deterministic by construction. SciCode's equivalent judge control disagrees
on 5 of 135 cells within one session, so determinism cannot be assumed of a
new benchmark.

\begin{table}[t]
\centering\footnotesize
\setlength{\tabcolsep}{4pt}
\caption{Per-cell representation contrasts: $\Delta U$ under one
representation minus $\Delta U$ under Raw on the same memory and task, with
task-clustered 95\% intervals. Only LongMemEval-V2 separates from zero. Every
SciCode and SWE-bench interval includes zero (SWE-bench's Summary$-$Raw only at
its upper bound), and doubling the SciCode cell
count separated no contrast. MDE is the minimum detectable contrast at 80\%
power and two-sided $\alpha = 0.05$, computed from each interval's width.
Raw's own with-versus-without $\Delta U$ is $\rvSciRaw$
$[\rvSciRawLo, \rvSciRawHi]$ on SciCode and $\rvSweRaw$ on SWE-bench, so on
these benchmarks no memory effect exists for a representation to preserve, and
the contrasts cannot test the ordering.}
\label{tab:crossbench}
\setlength{\tabcolsep}{3pt}
\begin{tabular}{@{}llrrrr@{}}
\toprule
contrast & benchmark & $n$ & mean & 95\% CI & MDE \\
\midrule
Summary$-$Raw
 & LongMemEval-V2 & \xbSRLmN & \xbSRLmM & [\xbSRLmLo, \xbSRLmHi] & \raMdeSRLm \\
 & SciCode        & \xbSRScN & \xbSRScM & [\xbSRScLo, \xbSRScHi] & \raMdeSRSc \\
 & SWE-bench      & \xbSRSwN & \xbSRSwM & [\xbSRSwLo, \xbSRSwHi] & \raMdeSRSw \\
\midrule
Structured$-$Raw
 & LongMemEval-V2 & \xbTRLmN & \xbTRLmM & [\xbTRLmLo, \xbTRLmHi] & \raMdeTRLm \\
 & SciCode        & \xbTRScN & \xbTRScM & [\xbTRScLo, \xbTRScHi] & \raMdeTRSc \\
 & SWE-bench      & \xbTRSwN & \xbTRSwM & [\xbTRSwLo, \xbTRSwHi] & \raMdeTRSw \\
\bottomrule
\end{tabular}
\end{table}

\section{Sign reversal and the decomposition}
\label{sec:negative}
\begin{figure}[t]
\centering
\includegraphics[width=\columnwidth]{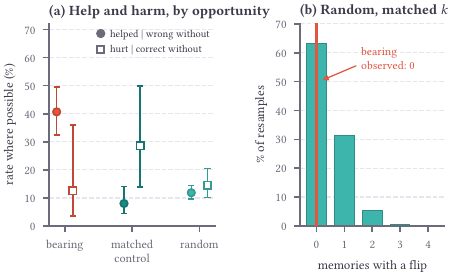}
\caption{Sign reversal on LongMemEval-V2, screened design (1{,}157 cells,
60 memories). \textbf{(a)}~Help rate among cells the reader got wrong without
the memory (filled) and harm rate among cells it got right (hollow), with
Wilson 95\% intervals; counts are in Table~\ref{tab:transitions}. Harm does not differ detectably across
strata; help does. \textbf{(b)}~Flips among the 35 memories with at least two
\textsc{bearing} cells when each memory's \textsc{random} cells are subsampled
to its own \textsc{bearing}-cell count (20{,}000 resamples); the line marks the
observed \textsc{bearing} count.}
\label{fig:reversal}
\Description{Left: dot plot of help and harm rates by stratum; help is about 41 percent for bearing against 8 and 12 percent elsewhere, harm overlaps across strata at 12 to 29 percent with wide intervals. Right: histogram of flip counts at matched cell counts, with zero flips in about 63 percent of resamples.}
\end{figure}

\paragraph{Why flip rates cannot be compared across strata.}
A memory can show a flip only if at least two of its cells move with opposite
signs, so the chance of a flip grows with the number of cells per memory. In
the screened design a memory has $\raCellsPerMemBear$ \textsc{bearing} cells
and $\raCellsPerMemRand$ \textsc{random} cells on average. The naive
contrast (0 of 35 \textsc{bearing} memories against 17 of 60 \textsc{random},
Fisher one-sided $p = 1.5\times10^{-4}$) therefore mostly reflects that
difference. To test this, we took each of the $\raMatchedN$ memories with at
least two \textsc{bearing} cells and drew, without replacement, as many of its
\textsc{random} cells as it has \textsc{bearing} cells, then recounted flips
(20{,}000 resamples). The random stratum averages $\raMatchedMean$ flips and
shows none in $\raMatchedPZero\%$ of resamples (Figure~\ref{fig:reversal}b), so
0 of 35 is what chance alone would most often produce. A Wilson bound on the \textsc{bearing} flip rate would also treat 35 memories
from 21 episodes as independent.
Matching each memory's random subsample on the no-memory outcome as well (so
both strata get the same opportunities to help and to hurt) leaves all
$\lfCondFlipMems$ memories eligible; the random stratum then shows a mean of
$\lfCondFlipMean$ flips and none in $\lfCondFlipPZero\%$ of resamples. We
therefore do not compare flip rates across strata.

\paragraph{Why the sign of moving cells is not a comparison either.}
A tempting alternative compares the share of negative cells among cells that
move: $\raNegBearK$ of $\raNegBearN$ for \textsc{bearing} against
$\raNegRandK$ of $\raNegRandN$ for \textsc{random}. That comparison is
confounded by baseline accuracy. A memory can only hurt where the reader was
correct without it, and \textsc{bearing} tasks have the lowest no-memory
accuracy, so they offer the fewest chances to hurt.

\paragraph{Transitions conditioned on the no-memory outcome.}
Reward is binary in the screened run: each stratum's mean equals
(helped $-$ hurt)$/n$ exactly. With Table~\ref*{M-tab:screened}'s both-arms-fail
rate this fixes the full transition table (Table~\ref{tab:transitions}).
Conditioned on opportunity, the strata do not differ detectably in harm,
and they differ sharply in help (Figure~\ref{fig:reversal}a). In the
\textsc{random} stratum help and harm rates are close, the signature of a
content-independent response to inserting an item, which we treat as the
perturbation floor (\main{sec:metrics}). The table's Wilson intervals treat cells as
independent. Resampling memories instead, the harm rates have intervals of
$[\lfHurtClBearLo, \lfHurtClBearHi]$ (\textsc{bearing}),
$[\lfHurtClCtrlLo, \lfHurtClCtrlHi]$ (control) and
$[\lfHurtClRandLo, \lfHurtClRandHi]$ (\textsc{random}), which overlap, and the
help rates $[\lfHelpClBearLo, \lfHelpClBearHi]$,
$[\lfHelpClCtrlLo, \lfHelpClCtrlHi]$ and $[\lfHelpClRandLo, \lfHelpClRandHi]$,
where \textsc{bearing}'s is disjoint from the others. Both
analyses here are exploratory and are produced by
\texttt{make\_reanalysis.py}, which first checks that it reproduces every flip
count and stratum mean in the paper.

\begin{table}[t]
\centering\footnotesize
\setlength{\tabcolsep}{2.6pt}
\caption{Screened run on LongMemEval-V2: transitions between the no-memory
and with-memory outcome (binary reward). Accuracy columns give the share of
cells answered correctly without and with the memory. Hurt is conditioned on
cells correct without the memory, help on cells wrong without it; Wilson 95\%
intervals in brackets.}
\label{tab:transitions}
\begin{tabular}{@{}lrrrrr@{}}
\toprule
stratum & $n$ & acc.\ w/o & acc.\ with & hurt $\mid$ correct & help $\mid$ wrong \\
\midrule
\textsc{bearing} & \spBearN & \raAccZeroBear & \raAccOneBear
  & \raHurtRateBear\ [\raHurtLoBear, \raHurtHiBear] & \raHelpRateBear\ [\raHelpLoBear, \raHelpHiBear] \\
control & \spCtrlN & \raAccZeroCtrl & \raAccOneCtrl
  & \raHurtRateCtrl\ [\raHurtLoCtrl, \raHurtHiCtrl] & \raHelpRateCtrl\ [\raHelpLoCtrl, \raHelpHiCtrl] \\
\textsc{random} & \spRandN & \raAccZeroRand & \raAccOneRand
  & \raHurtRateRand\ [\raHurtLoRand, \raHurtHiRand] & \raHelpRateRand\ [\raHelpLoRand, \raHelpHiRand] \\
\midrule
\multicolumn{6}{@{}l}{\emph{opportunities:} hurt $\raHurtOppBear$ / $\raHurtOppCtrl$ / $\raHurtOppRand$ cells;
help $\raHelpOppBear$ / $\raHelpOppCtrl$ / $\raHelpOppRand$ cells} \\
\bottomrule
\end{tabular}
\end{table}

\paragraph{The additive-null comparison.}
Fitting $\mu + a_m + b_t$ and simulating from it, the excess of the observed
reversal rate over the fitted additive model is negative in all three strata.
Two independent implementations agree on the sign and disagree on the
magnitude, so we quote only the sign.

\paragraph{Why the decomposition is not identifiable.}
The within-session execution floor is exactly zero: replicates issued 0--16
seconds apart agree on 1{,}157 of 1{,}157 cells. With no residual scale, the
fitted residual collapses and the task and interaction shares are not
separately identified, so we report none. Clean per-cell utilities and a
variance decomposition are in tension on a reader without execution noise.

\section{Data, code and prompt availability}
\label{sec:repro}
\label{sec:prompts}
\paragraph{Public dataset.} The Conditional Memory Utility Dataset is available on
\href{https://huggingface.co/datasets/AgentsSci/AAMAS2027-ConditionalMemory}{Hugging Face}.
The frozen \texttt{v0.6-}\allowbreak\texttt{length-control} snapshot contains 16,816 records:
16,535 from LongMemEval-V2, 154 from SciCode, and 127 from SWE-bench Verified.
A row describes a memory--task pair under a representation, reader, condition
and replicate, with the outcome under each arm, available output text and
cost metadata. Missing or unscorable measurements remain explicitly marked;
they must not be converted to zero rewards.

\paragraph{Version and reproducibility.} The data file is
\texttt{data/}\allowbreak\texttt{cells.jsonl}. The repository also provides a manifest with file checksums, schema and validation
code, and documentation of the included conditions and upstream sources.
Readers can download the data directly or pin the
\texttt{v0.6-}\allowbreak\texttt{length-control} revision with the Hugging Face client. The data
card contains a runnable download example and explains the validation warnings.
This snapshot was built on 2 October 2026 and does not contain every later
manuscript analysis; in particular, its condition inventory does not include
the fixed-total-context experiment. The public Hub package contains schema and
validation code, not the complete experiment, analysis and prompt bundle.
The length-control identity arm is an A/A control and must not be pooled with
Raw as an independent condition. Per-pair utility labels describe the recorded
evaluation conditions, rather than unconditional training targets.

\paragraph{Withheld and missing text.} The 154 SciCode records contain Raw
memories; their reader outputs were not retained. SciCode Summary and Structured
records are absent because their memory text was not persisted. The 127
SWE-bench records provide identifiers and derived measurements, with restricted
corpus and output text withheld under the release policy. These are distinct
from records missing because of a persistence gap. The data card documents the
sentinels and per-source reuse terms; access to the dataset does not grant a
new license to withheld upstream content.

\section{Extended limitations}
\label{sec:extlimits}

\paragraph{Leakage rates are lower bounds.}
Our leakage detector reuses the benchmark's phrase matcher, whose normalizer
strips punctuation without inserting a space. A gold phrase next to a quote
or an equals sign is glued to its neighbor and the match fails. Such leaks
are missed, and punctuation is common in accessibility trees, so every
leakage rate we report is a lower bound. A control test records the defect, so
an upstream fix would be detected. The 156
LLM-judged questions cannot be audited without an inference call and are
recorded as undecidable rather than clean. 69 multiple-choice questions with
gold answers of one to five characters cannot be audited structurally and are
never folded into a pooled rate.

\paragraph{The judge floor measures decoding.}
$F_{\mathrm{judge}}$ is zero wherever we measured it: over $\lfJudgeObs$
explicit re-scorings of $\lfJudgeUnits$ units in the earlier official arm, and
in every stratum of the screened run. Both estimators carry a positive control
(a planted disagreement makes them fire). A zero floor here reflects a
deterministic scorer and greedy decoding, not an unambiguous rubric. The
SciCode control in Section~\ref{sec:crossbenchtable}, which disagrees on 5 of
135 cells, is a different scorer on a different benchmark. We have no
cross-session floor for any arm and presume it to be non-zero.

\paragraph{Semantic type.}
We intended semantic memory type as a second axis, but in our representation
pilot the two labeling passes agreed on 3 of 8 segments at Cohen's
$\kappa = 0.000$: the structured writer's own pass called every segment
\texttt{episodic}, a degenerate marginal, so we report no semantic-type
breakdown. Separately, exactly one deterministically scorable
\emph{errors-gotchas} question exists in the benchmark and accounts for 865 of
6{,}920 web cells, so any per-stratum result there is a single-question result.

\paragraph{Writer failures and faithfulness.}
The structured writer failed on 3 of 60 segments, all by exhausting the token
budget mid-value rather than by malformed output, and on 0 of the 57 writes
that finished. An 8-segment pilot had put the failure rate at 0 of 8 and could
not have seen this, since it never reached the cap. Our faithfulness measures
are a screen, not a guarantee: a low score is strong evidence of a problem,
a high score is weak evidence of its absence, and they do not detect
recombination, omission, unsupported inference or negation.

\paragraph{Unrecorded quantities.}
Two quantities were not recorded during the runs and cannot be recovered: how
many gold-free picks contain the answer, and how the 12 position-probe cells
were chosen, which is why we treat that probe as a pilot.

\paragraph{Stated exclusions.}
17 of the 295 deterministically scored questions have no memory unit that
carries their gold answer under the segmentation answerability check, so their
cells can only yield $\Delta U \le 0$. The bearing screen, a looser containment
test, fires on no unit for $\soScrZero$ of them (Section~\ref{sec:sophia}).
23 of the 1{,}889 units (1.2\%) are their entire trajectory.

\section{Pre-registration record}
\label{sec:prereg}
Each pre-registered analysis was fixed in a pre-commitment document in the code
repository before the data it governs existed. The documents are released with
the code. To preserve anonymity we omit their commit identifiers here; the
release links each one to a commit and its timestamp, so its timing can be
checked on de-anonymization. Table~\ref{tab:prereg} lists every pre-registered analysis, its
anchor, and every deviation the run ledgers record. Pre-commitments were not edited after
the fact; where a result does not match one, the mismatch is reported.

\begin{table}[h]
\centering\footnotesize
\setlength{\tabcolsep}{3pt}
\caption{Pre-registered analyses, their anchors, and recorded deviations.}
\label{tab:prereg}
\begin{tabular}{@{}p{0.27\columnwidth}p{0.30\columnwidth}p{0.37\columnwidth}@{}}
\toprule
analysis & anchor & recorded deviation \\
\midrule
screened design and its reduction criterion & criterion fixed in code before the numbers existed & none recorded \\
reversal rate (headline) & pre-registration, section~5 & comparison with an additive null added later, as a diagnostic \\
variance decomposition & pre-registration, section~6 & not identifiable here (zero execution floor); in earlier arms the mixed model did not converge and the registered fallback was used \\
gold-free screens & gold-free screen pre-commitment & none recorded \\
length control & frozen at a recorded commit, before execution & \lfPrLcMismatch\ of \lfPrLcCompared\ descriptive figures do not re-derive: bearing quartile means by Raw length; none appears in the paper \\
second reader & written before any design-cell call & cross-session estimate superseded by a session-matched run \\
session-matched reader run & falsifiers committed before data & none recorded \\
writer\,$\times$\,reader factorial & factorial pre-commitment & one prose sign error ($-\tfrac12$ for $+\tfrac12$ interaction); replication check fails on sign only \\
SciCode scale-up & scale-up pre-commitment & none recorded \\
\bottomrule
\end{tabular}
\end{table}

\paragraph{The two deviations that bear on results.} The additive-null
comparison of reversal rates (\main{sec:r0}) was \emph{not} pre-registered;
the registered headline is the reversal rate itself, and the comparison is a
diagnostic added afterwards. In the writer\,$\times$\,reader factorial, the
registered criterion asked the re-measured reader effect to reproduce the
earlier one in sign and significance. Both estimates are near zero and span
zero (Summary $\lfPrWfSumOld$ then $\lfPrWfSumNew$, Structured
$\lfPrWfStrOld$ then $\lfPrWfStrNew$), so the significance verdict reproduces
and the sign does not; the two intervals contain each other's point estimates.
The factorial's pre-commitment also states that the same-family effect equals
$-\tfrac12$ of the interaction; the correct relation, which the analysis code
checks numerically, is $+\tfrac12$. Every quantity was computed as registered.

\section{Follow-up runs}
\label{sec:sophia}
Three follow-up runs test the main results under conditions the main design
held fixed; all are exploratory and used the same serving endpoint and reader as
the main runs. A generator script in the released code computes every number
below from the row files.

\paragraph{Second session, randomized position.} All $\soXoneCells$ bearing and
control cells, three replicates each, the memory's position redrawn per
replicate ($\soXoneCalls$ real calls, $\soXoneCacheHits$ from cache,
$\soXoneErr$ failed and excluded). Bearing minus control, paired within memory:
$\soXoneD$ $[\soXoneLo, \soXoneHi]$ over $\soXoneMems$ memories;
episode-clustered $[\soXoneEpLo, \soXoneEpHi]$ ($\soXoneEps$ episodes); two-way
(memory $\times$ task) $\soXoneTw$ $[\soXoneTwLo, \soXoneTwHi]$. Across the
three positions of a cell, the label changes in $\soPosChg$ of $\soPosCells$
cells ($\soPosChgPct\%$), and in $\soPosChgMv$ of the $\soPosMov$ cells that move
at least once ($\soPosChgMvPct\%$); position accounts for $\soPosVarPct\%$ of
the variance of $\Delta U$. Joined one-to-one to the published run, labels agree
in $\soStabSamePct\%$ of $\soStabSameN$ replicate-cells at the same position
($\kappa = \soStabSameK$) and in $\soStabDiffPct\%$ of $\soStabDiffN$ at a new
one ($\kappa = \soStabDiffK$).

\paragraph{Fixed total context.} $\soXtwoCells$ cells over $\soXtwoMems$
memories, two replicates. For each cell, $T$ is the base filler plus the Raw
memory; every other arm, and the no-memory arm, is padded after the base context
with bearing-disjoint filler to $T$ (asserted equal across arms), the memory
kept at its K2 position. \emph{Disclosed defect:} the no-memory cache key was
(trajectory, task, replicate), so two memories from one trajectory paired with
one task could share a no-memory response although their padding differs. The
headline numbers exclude all $\soXtwoCollCells$ such cells ($n=\soFN$ bearing
cells); including them ($n=\soFNAll$) changes no conclusion:

\begin{table}[!ht]
\centering\footnotesize
\setlength{\tabcolsep}{3pt}
\caption{Bearing-cell $\Delta U$ at a fixed total context (task-clustered
95\% intervals).}
\label{tab:fixedctx}
\begin{tabular}{@{}lll@{}}
\toprule
arm & excluding exposed cells & all cells \\
\midrule
Raw & \soFRaw\ [\soFRawLo, \soFRawHi] & \soFRawAll\ [\soFRawAllLo, \soFRawAllHi] \\
Summary & \soFSum\ [\soFSumLo, \soFSumHi] & \soFSumAll\ [\soFSumAllLo, \soFSumAllHi] \\
Structured & \soFStr\ [\soFStrLo, \soFStrHi] & \soFStrAll\ [\soFStrAllLo, \soFStrAllHi] \\
matched truncation & \soFTr\ [\soFTrLo, \soFTrHi] & \soFTrAll\ [\soFTrAllLo, \soFTrAllHi] \\
Summary $-$ Raw & \soFSumR\ [\soFSumRLo, \soFSumRHi] & \soFSumRAll\ [\soFSumRAllLo, \soFSumRAllHi] \\
Structured $-$ Raw & \soFStrR\ [\soFStrRLo, \soFStrRHi] & \soFStrRAll\ [\soFStrRAllLo, \soFStrRAllHi] \\
\bottomrule
\end{tabular}
\end{table}

The matched truncation is sensitive to the exclusion and supports no claim. The
cache key now includes the memory, so a re-run cannot repeat the defect.

\paragraph{Retrieval.} The reference labels are the screened design's (75 tasks, 266
bearing edges, reproduced and asserted). Each method ranks the task's whole
domain pool by the raw question; dense retrieval embeds 512-token chunks (a
unit scores its best chunk) or the first 2{,}000 tokens.

\begin{table}[!ht]
\centering\footnotesize
\setlength{\tabcolsep}{3pt}
\caption{Retrieval of bearing units (task-clustered 95\% intervals; chance
recall@10 is about \soChanceTenP). MRR is the mean reciprocal rank of the first
bearing unit; AUROC is the area under the ROC curve over bearing and
non-bearing units.}
\label{tab:retrieval}
\begin{tabular}{@{}lllll@{}}
\toprule
method & recall@1 & recall@10 & MRR & AUROC \\
\midrule
BM25 & \soRBmOne & \soRBmTen\ [\soRBmTenLo, \soRBmTenHi] & \soRBmMrr & \soRBmAuc \\
dense, prefix & \soRPreOne & \soRPreTen\ [\soRPreTenLo, \soRPreTenHi] & \soRPreMrr & \soRPreAuc \\
dense, chunks & \soRChkOne & \soRChkTen\ [\soRChkTenLo, \soRChkTenHi] & \soRChkMrr & \soRChkAuc \\
\bottomrule
\end{tabular}
\end{table}

Of the 295 deterministic questions, $\soScrZero$ fire on no unit, $\soScrDisc$
on one to eight, and $\soScrOver$ on more than eight. Without the cap, string
containment fires for all $\soNoCapMems$ memories, on $\soNoCapMin$ to
$\soNoCapMax$ tasks each.

\paragraph{Variance decomposition.} In the second-session run each cell has three
replicates at redrawn positions, which supplies the residual scale the main run
lacked. On the $\rvDecCells$ cells with all three replicates we fit the additive
model $\mu + a_m + b_t$ to cell means by least squares; its residual mean square
($\rvDecDf$ degrees of freedom) estimates the interaction variance plus the
within-cell variance divided by three, and the within-cell variance estimates the
residual. The shares are $\rvIntShare\%$ interaction, $\rvAddShare\%$ additive
memory and task effects, and $\rvEpsShare\%$ position with decoding; the
interaction $F$ test gives $F=\rvIntF$ ($p\rvIntP$). This is a method-of-moments
estimate on a sparse crossed design, reported as exploratory.

\paragraph{Power for harm.} With $16$ bearing cells able to show harm, against
the random stratum's rate of $\rvHarmRandPct\%$, the screened design detects
(80\% power, two-sided $\alpha=0.05$) only a bearing harm rate of
$\rvHarmMdePct\%$ or more; the null on harm is therefore weak evidence.

\paragraph{Containment inside the random-task stratum.} Memories in this stratum are real
same-domain pages. On the $\rvRandFireN$ random cells where the containment
detector fires (at any breadth) the memory helps in $\rvRandFireHelp\%$ of the
$\rvRandFireHelpN$ cells the reader got wrong without it, against
$\rvRandSilentHelp\%$ of $\rvRandSilentHelpN$ where it is silent; harm is
$\rvRandFireHurt\%$ and $\rvRandSilentHurt\%$. Part of the random stratum's help
is content, so its help rate bounds pure perturbation from above.

\paragraph{The gold-free screen over the full universe.} The model-based screen
was applied to all $\rvGfUniverse$ (task, unit) pairs: an exhaustive lexical
stage kept $\rvGfCand$ candidates, and the judge confirmed $\rvGfYes$ over
$\rvGfTasks$ tasks and $\rvGfMems$ memories, under the same cap of eight units
per task. A gold-free definition does not make bearing denser.

\paragraph{Sparsity across screen caps.} Recomputing the screen over the
deterministic tasks at four caps on the number of units a task may fire on:
at a cap of 4, $\rvCapFourTasks$ tasks and $\rvCapFourMems$ bearing memories;
at 8 (the paper's screen), $\rvCapEightTasks$ and $\rvCapEightMems$; at 16,
$\rvCapSixteenTasks$ and $\rvCapSixteenMems$; at 32, $\rvCapThirtyTwoTasks$ and
$\rvCapThirtyTwoMems$. At every cap the median bearing memory bears on exactly one
task, and $\rvCapShareLo$--$\rvCapShareHi\%$ bear on exactly one. Only removing the
cap entirely makes containment fire for every memory.

\paragraph{Which memories were sampled.} The 60 memories are every memory that
bears on two or more tasks and passes the size cap ($\rvSelMulti$ of the
corpus's $\rvPopMulti$) plus $\rvSelSingle$ single-task memories. The headline
contrast is $\rvHeadSingle$ $[\rvHeadSingleLo, \rvHeadSingleHi]$ on the
single-task memories and $\rvHeadMulti$ $[\rvHeadMultiLo, \rvHeadMultiHi]$ on the
multi-task ones; reweighted to the $\rvPopAll$ bearing memories of the corpus, of
which $\rvPopSinglePct\%$ bear on one task, it is $\rvHeadW$
$[\rvHeadWLo, \rvHeadWHi]$ (memory bootstrap within each group).

\section{Additional figures}
\label{sec:figs}
\begin{figure*}[p]
\centering
\includegraphics[width=\textwidth]{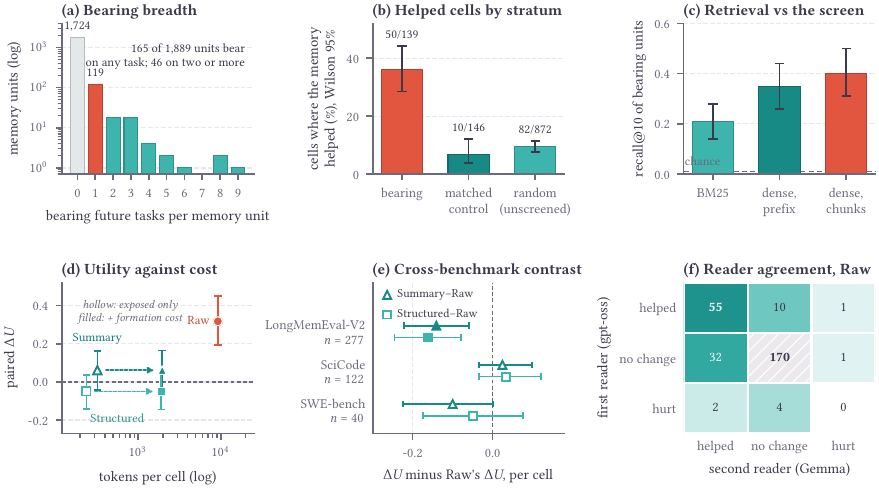}
\caption{Overview of further results on LongMemEval-V2 (a--d, f) and across
benchmarks (e). \textbf{(a)}~Future tasks each of the $\bearUnits$ memory units
bears on, over $\bearDetTasks$ discriminative tasks (log axis).
\textbf{(b)}~Share of cells where the memory helped, cell-level Wilson 95\%
intervals (memory-resampled ones in Section~\ref{sec:negative}), for
the same $\spMems$ memories in three strata. \textbf{(c)}~Share of bearing units a retriever ranks in its top ten of the
domain pool (75 tasks, task-clustered 95\% intervals); the dashed line is
chance. \textbf{(d)}~Paired
\textsc{bearing}$-$control effect per representation ($\frCells$ cells,
$\frMems$ memories, intervals over memories); hollow markers count exposed
tokens, filled ones add formation cost amortized over $\frReuse$ cells.
\textbf{(e)}~Per-cell $\Delta U$ minus Raw's, task-clustered intervals; filled
markers exclude zero. \textbf{(f)}~Two readers' labels on $\rgSmCells$ Raw cells
in one session; the hatched cell (zero under both) cannot disagree.}
\label{fig:overview}
\Description{Six panels: a log histogram of bearing tasks per memory unit peaking at one; bar chart of helped-cell rates by stratum with bearing highest; similarity sextile lines nearly flat; utility against tokens showing Raw above zero and compressed representations near zero; cross-benchmark contrast intervals, only LongMemEval-V2 excluding zero; a three by three reader agreement table.}
\end{figure*}

Figure~\ref{fig:overview} collects views referenced from the main paper: bearing
breadth, helped cells by stratum, retrieval, the representation contrasts with
their token costs, the cross-benchmark contrasts, and the two readers'
agreement. Figure~\ref{fig:lenctl} (Section~\ref{sec:lengthcontrol}) details the
length-control run.

\end{document}